%% file: main.tex
\documentclass[conference]{IEEEtran}
\IEEEoverridecommandlockouts %

\usepackage{times}
\usepackage{graphicx}

\usepackage{booktabs} %
\usepackage{multirow} %

\usepackage{amssymb}
\usepackage{xcolor}
\usepackage{amsmath}

\usepackage{algorithm}
\usepackage{algpseudocode}
\usepackage{array}
\usepackage{makecell}
\usepackage{colortbl}
\usepackage{textcomp}
\usepackage{placeins}
\usepackage{gensymb}
\usepackage{siunitx}
\usepackage{subcaption}
\usepackage[numbers]{natbib}
\usepackage{multicol}
\usepackage[bookmarks=true]{hyperref}
\usepackage[capitalise]{cleveref}

\newcommand{\ie}{\textit{i}.\textit{e}., }
\newcommand{\eg}{\textit{e}.\textit{g}., }

\newcolumntype{C}[1]{>{\centering\arraybackslash}p{#1}}

\definecolor{ourscolor}{HTML}{FCE5CD}

\usepackage{etoolbox}
\usepackage{caption}

\IEEEpeerreviewmaketitle  

\makeatletter
    \apptocmd{\@maketitle}{
        \centering
        \vspace{6pt}
        \captionsetup{type=figure}
        \includegraphics[width=\textwidth]{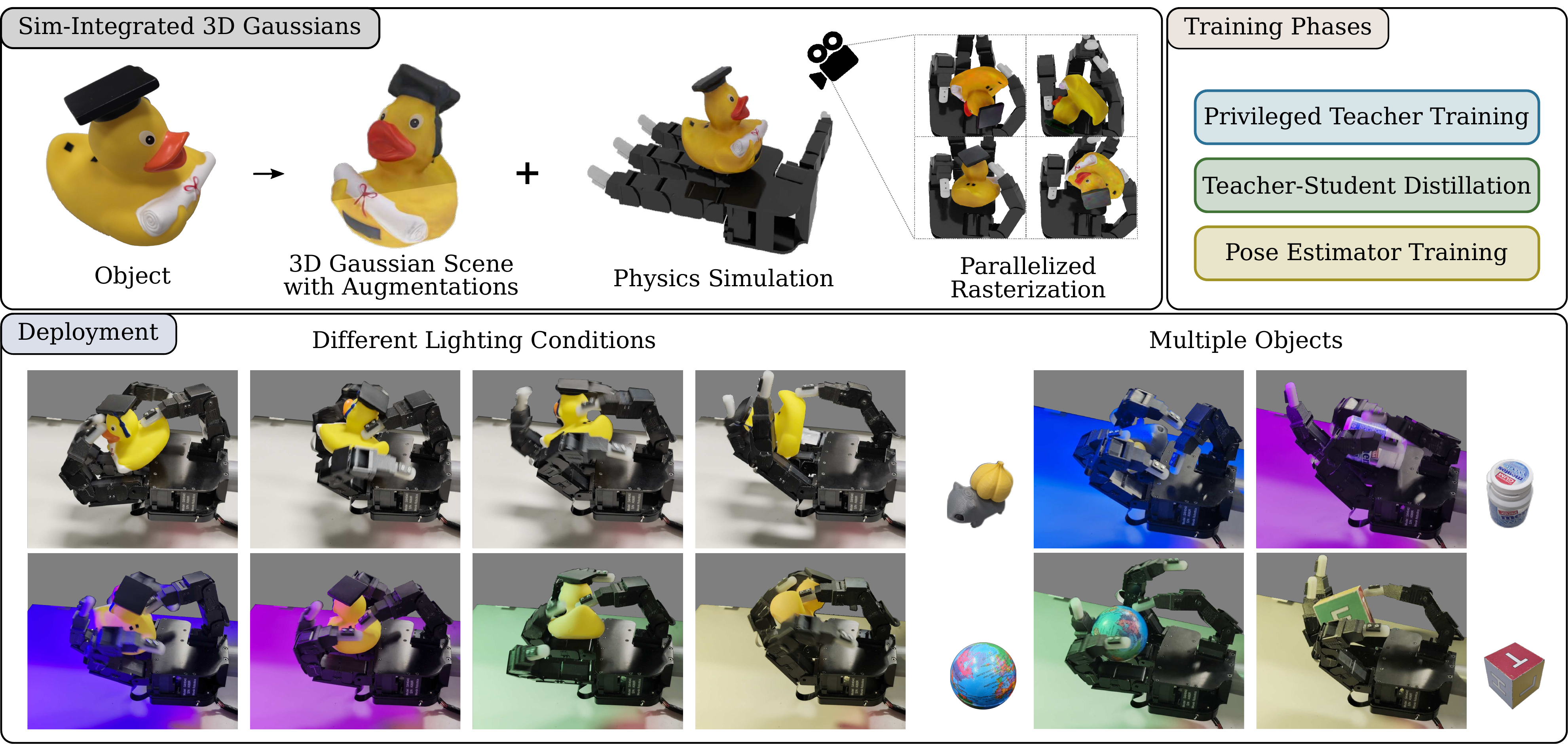}
        \captionof{figure}{We introduce a pipeline for training vision-based policies in simulation using 3D Gaussian Splatting. We successfully deploy these policies to the real world using a single monocular RGB camera, demonstrating robust in-hand reorientation of complex object geometries even under adversarial lighting conditions.}
        \label{fig:teaser}
        \vspace{-6pt}
    }{}{}%
\makeatother

\renewcommand{\baselinestretch}{0.995}

\begin{document}
  
\title{ViserDex: \textbf{Vi}sual \textbf{S}im-to-\textbf{R}eal for Robust \\ \textbf{Dex}terous In-hand Reorientation}

\author{
    \authorblockN{
        Arjun Bhardwaj$^{1}$,
        Maximum Wilder-Smith$^{1}$,
        Mayank Mittal$^{1,2}$, 
        Vaishakh Patil$^{1}$ and
        Marco Hutter$^{1}$
    }
    \authorblockA{$^{1}$ETH Zurich, $^{2}$NVIDIA\\Email: \texttt{\{abhardwaj, mwilder, mittalma, patilv, mahutter\}@ethz.ch}}
}

\maketitle
\setcounter{figure}{1}

\begingroup
\let\clearpage\relax

\input{chapters/0_abstract_clean}

\IEEEpeerreviewmaketitle

\input{chapters/1_introduction}

\input{chapters/2_related_works}

\input{chapters/3_methodology}

\input{chapters/4_results}
\input{chapters/5_conclusion}

\section*{Acknowledgments}
This work was funded by ETH Zurich (research grant \mbox{no. 22-2 ETH-47}), Swiss National Science Foundation (NCCR Automation grant \mbox{no. 51NF40\_225155}), Swiss Federal Railways (SBB) via ETH Mobility Initiative, ETHAR and by grants from NVIDIA. The authors also acknowledge the use of NVIDIA RTX 6000 Ada Generation GPUs, which facilitated this research.

The authors would like to thank René Zurbrügg for insightful discussions regarding pose estimator training and system identification, Pascal Roth for the initial integration of Gaussian splatting rendering with IsaacLab, and Elena Krasnova for assistance with the robotic hardware.

\bibliographystyle{plainnat}
\bibliography{references}

\clearpage
\thispagestyle{empty} %
\newpage
\clearpage

\begin{appendices} %
\input{chapters/A_appendix}
\end{appendices}

\endgroup

\end{document}

%% file: chapters/0_abstract_clean.tex
\begin{abstract}

In-hand object reorientation requires precise estimation of the object pose to handle complex task dynamics.
While RGB sensing offers rich semantic cues for pose tracking, existing solutions rely on multi-camera setups or costly ray tracing.
We present a sim-to-real framework for monocular RGB in-hand reorientation that integrates 3D Gaussian Splatting (3DGS) to bridge the visual sim-to-real gap.
Our key insight is performing domain randomization in the Gaussian representation space: by applying physically consistent, pre-rendering augmentations to 3D Gaussians, we generate photorealistic, randomized visual data for object pose estimation.
The manipulation policy is trained using curriculum-based reinforcement learning with teacher–student distillation, enabling efficient learning of complex behaviors.
Importantly, both perception and control models can be trained independently on consumer-grade hardware, eliminating the need for large compute clusters.
Experiments show that the pose estimator trained with 3DGS data outperforms those trained using conventional rendering data in challenging visual environments.
We validate the system on a physical multi-fingered hand equipped with an RGB camera, demonstrating robust reorientation of five diverse objects even under challenging lighting conditions.
Our results highlight Gaussian splatting as a practical path for RGB-only dexterous manipulation. 
For videos of the hardware deployments and additional supplementary materials, please refer to the project website: \url{https://rffr.leggedrobotics.com/works/viserdex/}.

\end{abstract}

%% file: chapters/1_introduction.tex
\section{Introduction}

Robotic dexterity requires not only grasping objects but also reorienting them within the hand into precise, functional poses. Deep Reinforcement Learning (DRL) has shown promise in acquiring such skills~\cite{andrychowicz2020learning,handa2023dextreme}. However, existing methods often succeed only with visually simple objects, such as colored cubes, and struggle with realistic textures, complex shapes, and varied appearances. A major challenge is the perception–control gap: rapid in-hand motions create severe self-occlusions, making accurate object pose estimation from real sensors extremely difficult.
Alternative sensing modalities, such as tactile arrays~\cite{pitz2024learning,touch-dexterity}, depth cameras~\cite{chen2023visual}, or multi-view rigs~\cite{handa2023dextreme}, offer partial solutions but introduce instrumentation overhead, calibration complexity, or limited scalability. Consequently, sim-to-real dexterous control has often avoided relying primarily on monocular RGB camera observations, resulting in a perception–robustness bottleneck that constrains current approaches.

A fundamental challenge in RGB-based manipulation lies in the simulation pipeline. Achieving the photorealism required for robust sim-to-real transfer via standard mesh-based rendering is computationally intractable for high-throughput RL training. Existing methods~\cite{singh2024dextrah, singh2025synthetica} attempt to address this gap using high-fidelity visual simulation; however, generating sufficient visual diversity over days of training demands massive compute clusters, even for simple objects~\cite{handa2023dextreme}. 
Explicit scene representations, particularly 3D Gaussian Splatting (3DGS)~\cite{kerbl20233dgs}, enable real-time, photorealistic rendering that outperforms traditional mesh rasterization. Its compact and flexible scene representation allows efficient manipulation of scenes, making it ideal for RL tasks that require large-scale visual diversity. Despite these advantages, standard 3DGS is limited to static scenes and entangles illumination with geometry~\cite{kerbl20233dgs}. This prevents independent manipulation of lighting and material properties, which is a prerequisite for Domain Randomization (DR) that facilitates robust sim-to-real transfer. In this work, we investigate strategies to overcome these limitations to generate diverse, dynamic visual data for robust object pose estimation during dexterous manipulation.

This paper proposes a monocular RGB-based training and deployment pipeline for robust in-hand reorientation of complex objects. The system is decomposed into two components: object pose estimation from RGB images as geometric keypoints, and an RL control policy that reorients the object to a desired goal pose. We integrate 3DGS directly into the simulation loop, overcoming the limitations of standard mesh-based rendering for high-throughput simulation-based training. Our key contribution is a suite of \textit{pre-rasterization augmentations} for Gaussian scenes. These augmentations generate consistent, diverse visual data and relax the static scene assumption of vanilla 3DGS. Furthermore, we simplify the RL process by replacing the extensive DR schemes used in previous works~\cite{akkaya2019solving, handa2023dextreme} with a performance-based curriculum and a student-teacher distillation framework, substantially improving training efficiency. Notably, our pipeline enables learning complex manipulation behaviors using only a single consumer-grade GPU. We validate this approach through zero-shot sim-to-real transfer on a 16-DoF Allegro Hand with a monocular RGB camera. Our approach demonstrates robust performance across five objects under both nominal and adversarial lighting conditions (shown in~\cref{fig:teaser}), achieving over 25 consecutive successful reorientations on average.

%% file: chapters/2_related_works.tex
\section{Related Works}

\subsection{Sim-to-Real RL for In-Hand Manipulation}

In-hand manipulation tasks in sim-to-real RL can be categorized into two primitives: continuous in-hand rotation and goal-conditioned reorientation. Continuous rotation, where the objective is to spin an object around a canonical axis, has been demonstrated using proprioceptive and tactile feedback~\cite{qi2023general, yang2024anyrotate, touch-dexterity}, as well as through repetitive open-loop finger gaits~\cite{bhatt2022surprisingly}.

Goal-conditioned reorientation, in contrast, requires precise object state tracking and geometric reasoning to repose the object to the target configuration. \citet{pitz2024learning} propose a tactile-based object state estimator. While tactile sensing captures local geometry, it lacks a global reference frame, making their method susceptible to drift over long horizons and unable to resolve fine-grained features. Depth-based approaches~\cite{chen2023visual} provide geometric structure but miss the semantic texture information needed to disambiguate the orientation of symmetric or visually complex objects.

RGB vision provides dense semantic feedback useful for robust in-hand reorientation. Prior works~\cite{andrychowicz2020learning, handa2023dextreme} have applied this modality to simple objects, such as colored cubes, but typically rely on multi-camera setups to handle occlusions. These methods also use computationally expensive Automatic Domain Randomization (ADR)~\cite{akkaya2019solving} to bridge the sim-to-real gap, requiring large-scale compute clusters. While recent efforts improve training efficiency through student-teacher distillation~\cite{singh2024dextrah}, generating photorealistic visual data remains a bottleneck.
Developing a monocular RGB-based framework that is both computationally efficient and generalizes to complex, real-world objects remains an open challenge.

\subsection{3D Gaussian Splatting for Robotic Manipulation}

3D Gaussian Splatting (3DGS)~\cite{kerbl20233dgs} was developed for fast novel-view synthesis, generating photorealistic images from unseen viewpoints. Its rapid, high-fidelity reconstruction has been adopted in robotics for SLAM~\cite{matsuki2024gsslam,hhuang2024photoslam}, teleoperation~\cite{wildersmith2024rfteleoperation, lee2026humanintheloop}, and planning~\cite{chen2024splatnav,michauxisaacson2024splanning}. The ability to easily capture and reconstruct arbitrary objects makes 3DGS a promising tool for reducing sim-to-real gaps in deploying visual policies.

Methods such as SplatSim~\cite{qureshi2024splatsimzeroshotsim2realtransfer} leverage 3DGS to produce higher-fidelity observations than mesh-based renders, improving realism and reducing sim-to-real discrepancies in manipulation tasks. However, traditional 3DGS scenes are optimized for single objects or static scenes, which limits domain randomization on them. GSRL~\cite{wang2024rlwithgeneralizedgs} addresses this by training a network to accelerate Gaussian generation across multiple scenes, and RL-GSBridge~\cite{wu2025rl-gsbridge} and RoboGSim~\cite{li2024robogsimreal2sim2realroboticgaussian} combine physics-compatible meshes with high-quality Gaussian rendering for visual RL. RoboGSim further expands the training domain with a Scene Composer that randomizes objects, backgrounds, and viewpoints.
Our work extends 3DGS by integrating it directly into a high-throughput simulation loop, combined with pre-rasterization augmentations. This approach addresses both visual realism and domain diversity challenges, paving the way for efficient sim-to-real transfer of dexterous manipulation policies.

%% file: chapters/3_methodology.tex
\section{Method}
\label{sec:method}

\begin{figure*}[h]
    \centering
    \includegraphics[width=\textwidth]{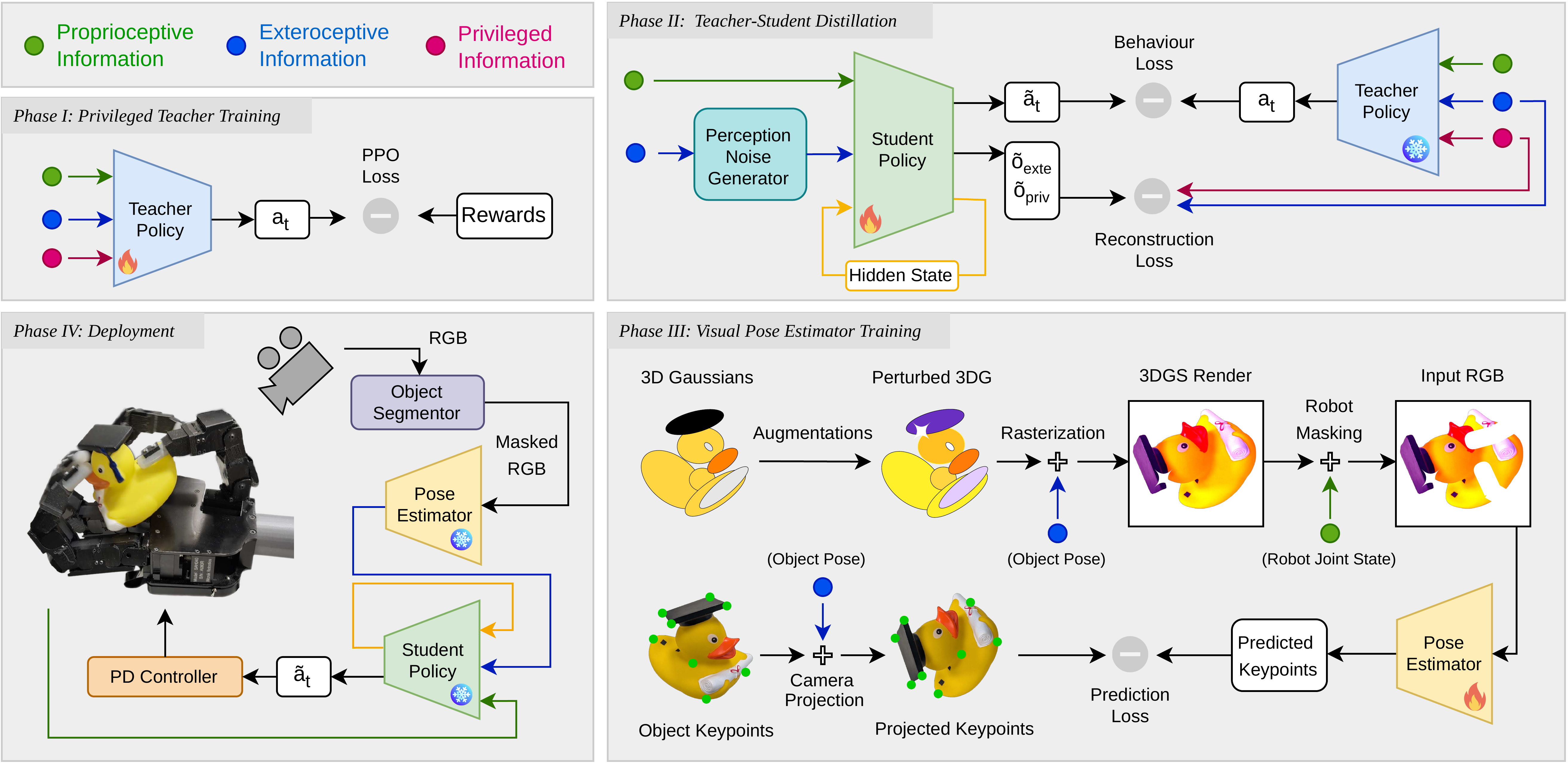}
    \caption{Overview of our sim-to-real in-hand reorientation pipeline. We first train a teacher policy in simulation with full state access, then distill it into a recurrent student policy that operates from noisy observations. A monocular RGB pose estimator trained with 3D Gaussian Splatting data provides object poses to the student policy, enabling goal-conditioned dexterous manipulation on a real multi-fingered hand.}
    \label{fig:pipeline_overview}
    \vspace{-16pt}
\end{figure*}

We consider the problem of continuous, goal-conditioned in-hand object reorientation using a multi-fingered robotic hand observed only through a monocular RGB camera. From a control perspective, this task constitutes a Partially Observable Markov Decision Process (POMDP) with severe perceptual challenges. During manipulation, the object undergoes rapid motion, frequent self-occlusions by the fingers, and significant motion blur, making direct state estimation from RGB images highly unreliable. At the same time, learning the underlying dexterous manipulation skills requires accurate reasoning about object geometry and contact dynamics, which is difficult to acquire from raw visual input alone.

A naive end-to-end approach that learns control directly from images must simultaneously solve three difficult problems: learn dexterous motor skills, infer hidden object state from partial observations, and overcome the visual sim-to-real gap. Jointly optimizing these objectives with reinforcement learning is challenging and sample inefficient. Instead, we follow a principled decomposition that addresses each challenge in a separate phase:
\begin{enumerate}
    \item \textbf{Teacher Training using RL:} We first train a teacher policy in simulation with full state access using reinforcement learning. This stage focuses solely on acquiring the geometric and contact-rich manipulation skills required for reorientation.
    \item \textbf{Student Distillation:} We distill the teacher into a recurrent student policy that learns to infer the underlying system state from noisy, real-world observations.
    \item \textbf{Visual Pose Estimator Training:} We train a monocular RGB pose estimator using synthetic images rendered from a 3D Gaussian Splatting representation of the object. By performing domain randomization directly in the Gaussian space prior to rendering, we generate photorealistic and diverse training data that significantly reduces the visual sim-to-real gap.
\end{enumerate}
At deployment, we predict object pose estimates from RGB images, which are fed to the recurrent student policy to produce goal-conditioned actions on the real robot. This modular design enables efficient training on consumer-grade hardware. An overview of the pipeline is shown in~\cref{fig:pipeline_overview}.

\subsection{Teacher Training using Reinforcement Learning}

\subsubsection{MDP Formulation} We formulate the reorientation task as a goal-conditioned MDP, and train a \mbox{policy $\pi_\theta(a_t | o_t, g_t)$} to rotate the object to a target orientation $g_t \in \mathrm{SO}(3)$ using PPO~\cite{Schulman2017ProximalPO}. The policy commands the robot's joint position targets $a_t \in \mathbb{R}^{16}$. The agent receives a dense reward for aligning the object with the goal, a sparse success bonus, and penalties that encourage smooth actions. Upon reaching a goal, a new target is sampled. Episodes terminate if the object is dropped or if no goals are achieved within a specified time window. Additional details are provided in the Appendix.

\subsubsection{Observations}
\label{sec:observation_groups}
We divide the observation space $\mathcal{O}$ into three groups: \emph{proprioceptive}, \emph{exteroceptive}, and \emph{privileged}. Proprioceptive observations $\mathcal{O}_{\text{prop}}$ consist of robot's joint positions, action history from the last four steps, the current object goal, and the remaining episode time. Exteroceptive-derived  observations~$\mathcal{O}_{\text{exte}}$ provide the object's current pose in the hand frame and its orientation relative to the goal.
Privileged observations $\mathcal{O}_{\text{priv}}$ is available only to the teacher policy and contain ground-truth state information, including the object's velocity, robot's fingertip contact forces, and randomized physical properties (\eg object mass and scale). %

\subsubsection{Performance-based Curriculum} 
\label{sec:curriculum_approach}

Prior work on in-hand reorientation~\cite{akkaya2019solving,handa2023dextreme} uses ADR over many environment parameters, which is computationally expensive. Instead, we propose a lightweight, performance-driven curriculum~\cite{10.5555/3455716.3455897} that increases the task complexity according to the agent's average consecutive success count. The curriculum has three complementary components. First, we gradually increase the regularization penalties, allowing the policy to prioritize task completion before refining smoothness and efficiency. Second, we incrementally increase the random action latency to prepare the agent for asynchronous delays on real hardware. Finally, we also progressively narrow the allowed time window between consecutive successes, encouraging more efficient object reorientation. Each curriculum component scales with the moving average of consecutive successes over all the environments. Together, these mechanisms improve sample efficiency and stabilize training, shown later in~\cref{sec:rl-policy-results}.

\subsection{Student Training using Distillation}

The RL policy from the previous phase has access to privileged signals, which are unavailable during deployment. We therefore distill it into a student policy which receives noisy proprioception~${o}_{\text{prop}}^{\text{noisy}}$ and noisy exteroceptive ~${o}_{\text{exte}}^{\text{noisy}}$ observations. To handle partial observability, we parameterize the student as a recurrent network with a belief encoder~\cite{takahiroperceptivelocomotion}, allowing it to implicitly infer the system state.

\subsubsection{Perception Noise Generator}

Inspired by~\cite{handa2023dextreme}, we corrupt simulated object pose observations with four perturbations. We apply temporal downsampling to simulate low frame rates, stochastic jitter to model variable latency, systematic bias for calibration errors, and inject random poses for occasional tracking failures. This realistic noise improves student policy robustness and facilitates effective sim-to-real transfer.

\subsubsection{Student Policy Architecture}

The student policy uses a \emph{belief encoder-decoder} network architecture, previously applied in perceptive locomotion~\cite{takahiroperceptivelocomotion}. At each timestep, the recurrent encoder updates a latent belief state $z = f_{\phi}({o}_{\text{prop}}^{\text{noisy}},{o}_{\text{exte}}^{\text{noisy}})$. During training, the belief decoder network reconstructs the teacher's observations $(\tilde{o}_{\text{exte}}, \tilde{o}_{\text{priv}}) = h_\psi(z, {o}_{\text{exte}}^{\text{noisy}})$. The encoder-decoder is trained with a reconstruction loss $\mathcal{L}_{\text{recon}}(\phi, \psi)$, which penalizes errors in reconstructing the privileged and exteroceptive information. This encourages the latent $z$ to capture the structure necessary to combine noisy inputs and compensate for partial observability.

The control head outputs the actions~$\tilde{a} = g_{\rho}(z, {o}_{\text{prop}}^{\text{noisy}}, {o}_{\text{exte}}^{\text{noisy}})$, supervised with a behavior cloning loss $\mathcal{L}_{\text{BC}}(\phi, \rho)$, that minimizes the L2 distance between the student and teacher actions. We train the student policy end-to-end using a composite loss function $\mathcal{L} = \mathcal{L}_{\text{BC}} + \lambda \mathcal{L}_{\text{recon}}$ and employ an online variant of DAgger~\cite{ross2011reduction} detailed in the Appendix.

\subsection{Visual Object Representation and Augmentations}

The student policy’s exteroceptive input requires object pose estimates from RGB images. A key challenge in training a robust perception model for this task is generating diverse visual data that handles lighting variations and heavy occlusions. To avoid the computational overhead of ray-tracing in simulators, we integrate Gaussian Splatting rasterization into the simulation loop.

\subsubsection{3D Gaussian Object Representation}
We represent object geometry and appearance using 3DGS~\cite{kerbl20233dgs}. The object is represented by a set of 3D Gaussians, each characterized by a position, covariance, opacity, and spherical harmonic (SH) coefficients~\cite{yufridovichkeil2021plenoxels}. To capture view-dependent effects, the color $c(\mathbf{d})$ along the viewing direction $\mathbf{d}$ is computed by spherical harmonics up to degree $L=3$:
\begin{equation}
    c(\mathbf{d}) = \text{Sigmoid}\left(\sum_{\ell=0}^{L} \sum_{m=-\ell}^{\ell} k_{\ell}^m Y_{\ell}^m(\mathbf{d})\right)
    \label{eq:sh_color}
\end{equation}
\noindent where $k_{\ell}^m$ are the learned coefficients. The $0^{th}$-order coefficients (SH0) capture view-independent Lambertian base colors, while the higher-order coefficients (SHN) encode high-frequency specular effects.

\subsubsection{Simulation-Integrated GS Rendering}
\label{rendering}

During in-hand manipulation, the camera is fixed while the object moves due to robot interactions. To render the moving object with 3DGS, we apply the inverse of the object's transform to the camera and produce RGB and depth $D_{\text{splat}}$. This transformation keeps the scene as static, satisfying the assumptions of vanilla 3DGS, while producing images that reflect changing object poses.
However, the object-centric rendering ignores occlusions from the robot's fingers.
To restore physical consistency, we generate a depth map of the hand $D_{\text{phys}}$ from the same viewpoint using a low-fidelity depth raycaster within the physics simulator. We mask out pixels in the RGB image where the hand is in the front ($D_{\text{phys}} < D_{\text{splat}}$). This aligns visual observations with the simulation state without full-scene ray-tracing.

\begin{figure}
    \centering

    \includegraphics[width=0.85\linewidth]{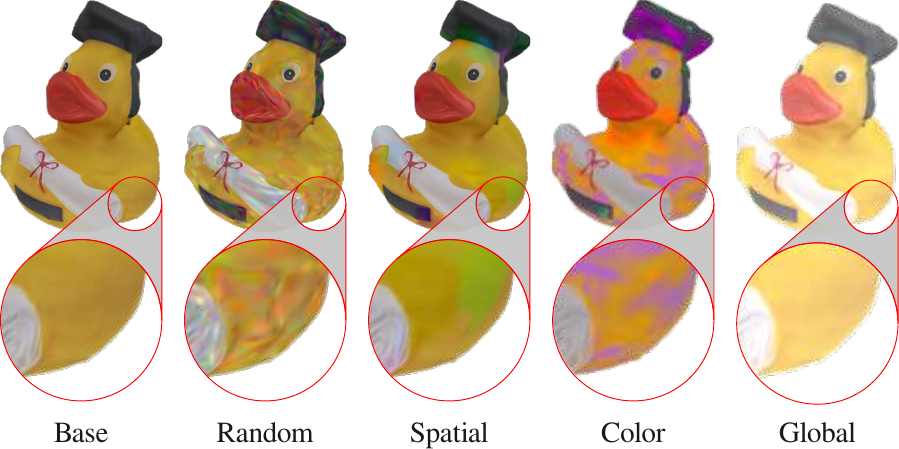}

    \caption{Pre-rasterization augmentation examples. Visualizations of the proposed SH-based perturbations applied to clustered Gaussians, producing structured variations in color, reflectance, and spatial appearance without ray-tracing.}
    \label{fig:augmentations}
    \vspace{-10pt}
\end{figure}

\subsubsection{Pre-Rasterization Augmentations}
\label{sec:augmentations}

Strategies for generating diverse visual data typically fall into two categories. First, \textit{post-process image augmentations} (\eg color jitter, brightness), which are computationally cheap but apply global 2D transformations that disregard 3D geometry. Second, \textit{scene parameter randomizations} (domain randomization)~\cite{singh2025synthetica} alter lighting and materials during rendering but require computationally expensive ray-tracing.

We propose a hybrid approach termed \textit{pre-rasterization augmentation}, which leverages the explicit nature of the 3D Gaussian representation. We operate directly on the Gaussian attributes, specifically the Spherical Harmonic (SH) coefficients, before rasterization. This provides fine-grained control over scene appearance without the cost of full scene ray-tracing. Na\"ively randomizing individual Gaussians, however, breaks photometric consistency, producing high-frequency noise rather than realistic lighting variations.
To generate diverse yet plausible data, we exploit the fact that lighting and material changes are inherently structured, typically affecting spatially proximal regions or specific materials uniformly. We therefore cluster Gaussians based on geometric or photometric correlations and perturb each cluster as a group. These perturbations include additive and scaling noise on the base color (SH0) and specular components in higher-order SH coefficients (SHN).
They are summarized as follows:

\begin{table}[t]
    \centering
    \caption{Pre-Rasterization Augmentation Parameters}
    \label{tab:splat_augmentations}
    \resizebox{\columnwidth}{!}{%
    \begin{tabular}{l l c c c}
    \toprule
    \textbf{Augmentation} & \textbf{Targets} & \textbf{Probability} & \textbf{Fraction} & \textbf{Range} \\
    \midrule
    \multicolumn{5}{l}{\textbf{\textsc{Random Noise}}} \\
    \addlinespace[0.3em]
    Additive    & SH0, SHN & 0.2 & 1.0 & $[-0.1, 0.1]$ \\
    Scaling     & SH0, SHN & 0.2 & 1.0  & $[0.8, 1.2]$ \\
    \addlinespace[0.6em]
    
    \multicolumn{5}{l}{\textbf{\textsc{Spatial Cluster}}} \\
    \addlinespace[0.3em]
    Additive    & SH0, SHN & 0.8 & 0.10 & $[-0.1, 0.1]$ \\
    Scaling     & SH0, SHN & 0.8 & 0.20 & $[0.9, 1.1]$ \\
    \addlinespace[0.6em]
    
    \multicolumn{5}{l}{\textbf{\textsc{Color Cluster}}} \\
    \addlinespace[0.3em]
    Additive    & SH0      & 0.8 & 0.10 & $[-0.2, 0.2]$ \\
    Additive    & SHN      & 0.8 & 0.10 & $[-0.1, 0.1]$ \\
    Scaling     & SH0, SHN & 0.8 & 0.10 & $[0.6, 1.4]$ \\
    \addlinespace[0.6em]
    
    \multicolumn{5}{l}{\textbf{\textsc{Global Shift}}} \\
    \addlinespace[0.3em]
    Additive    & SHN      & 0.2 & 1.0 & $[-0.1, 0.1]$ \\
    Scaling     & SH0, SHN & 0.2 & 1.0 & $[0.6, 1.4]$ \\
    Uniform Additive    & SH0, SHN & 0.8 & 1.0   & $[-0.2, 0.2]$ \\
    Uniform Scaling       & SH0      & 0.8 & 1.0 & $[0.9, 1.4]$ \\
    \bottomrule
    \end{tabular}
    }
    \vspace{-10pt}
\end{table}

\begin{itemize}
    \item \textbf{Random Noise Group:} We apply random perturbations to each Gaussian independently. This unstructured randomization effectively simulates high-frequency sensor noise, pixel-level artifacts, and minor mesh imperfections.
    
    \item \textbf{Spatial Cluster Group:} Real-world variations such as shadows, damage, and marks are often localized in a small region on the object. To mimic this effect, we cluster Gaussians by spatial location into 64 clusters using k-means. Perturbing these spatially contiguous clusters simulates local inconsistencies and patch-level noise.
    
    \item \textbf{Color Cluster Group:} Objects are often composed of distinct materials that interact differently with light. Assuming that these material properties correlate with diffuse color, we cluster Gaussians based on their $0^{th}$ order spherical harmonic (SH0) into 32 clusters. Perturbing these clusters simulates material-specific shifts, such as albedo modifications or reflectance changes, across photometrically similar regions.
    
    \item \textbf{Global Shift Group:} To simulate macro-level environmental changes, we treat the entire scene of Gaussians as a single cluster. We apply global shifts and scaling to the color attributes, where noise is sampled either independently per dimension or as a single value (Uniform) for the entire vector. These perturbations effectively replicate environmental variations such as ambient brightness, color temperature, camera exposure, and saturation.
\end{itemize}

Each Gaussian scene is preprocessed to identify the cluster indices for these strategies. During visual data generation, we sequentially apply stochastic variations to the SH coefficients, allowing different visual perturbations to compound. This produces a diverse set of scene appearances from a single static representation, as shown in~\cref{fig:augmentations}.~\cref{tab:splat_augmentations} summarizes the augmentation types and parameters. Each augmentation is applied with a specified probability, and only a fraction of the clusters are perturbed at a time. Despite their distinct semantic targets (\eg spatial vs. color), the procedural logic for applying any augmentation layer is consistent.

\subsection{Visual Object Pose Estimator Training}

We train a keypoint-based pose estimator to recover the object pose from RGB images, thereby providing the exteroceptive input required by the student policy during real-world deployment. The training dataset is generated by rolling out the expert teacher policy within the simulation and rendering the RGB images through our 3DGS pipeline with pre-rasterization augmentation strategies. This process results in a large-scale, annotated dataset with ground-truth object poses.
 To improve robustness to real-world sensor imperfections, we apply random ISO noise and motion blur to the images during training.
 
The pose estimator uses a ResNet-34~\cite{He2015DeepRL} backbone, initialized with ImageNet-pretrained weights. The network is trained to regress a set of nine keypoints, corresponding to the eight object-specific points plus the geometric centroid. For each keypoint, the network predicts the normalized 2.5D coordinates $(u, v, d)$, where $(u, v)$ represent the normalized pixel coordinates in the image plane and $d$ represents the metric depth. The predicted 2.5D keypoints can be resolved to a 6D object pose via the Rigid Procrustes algorithm~\cite{ten_Berge_2006}.

\subsection{Experimental Setup}

\begin{figure}
    \centering
    \includegraphics[width=0.975\linewidth]{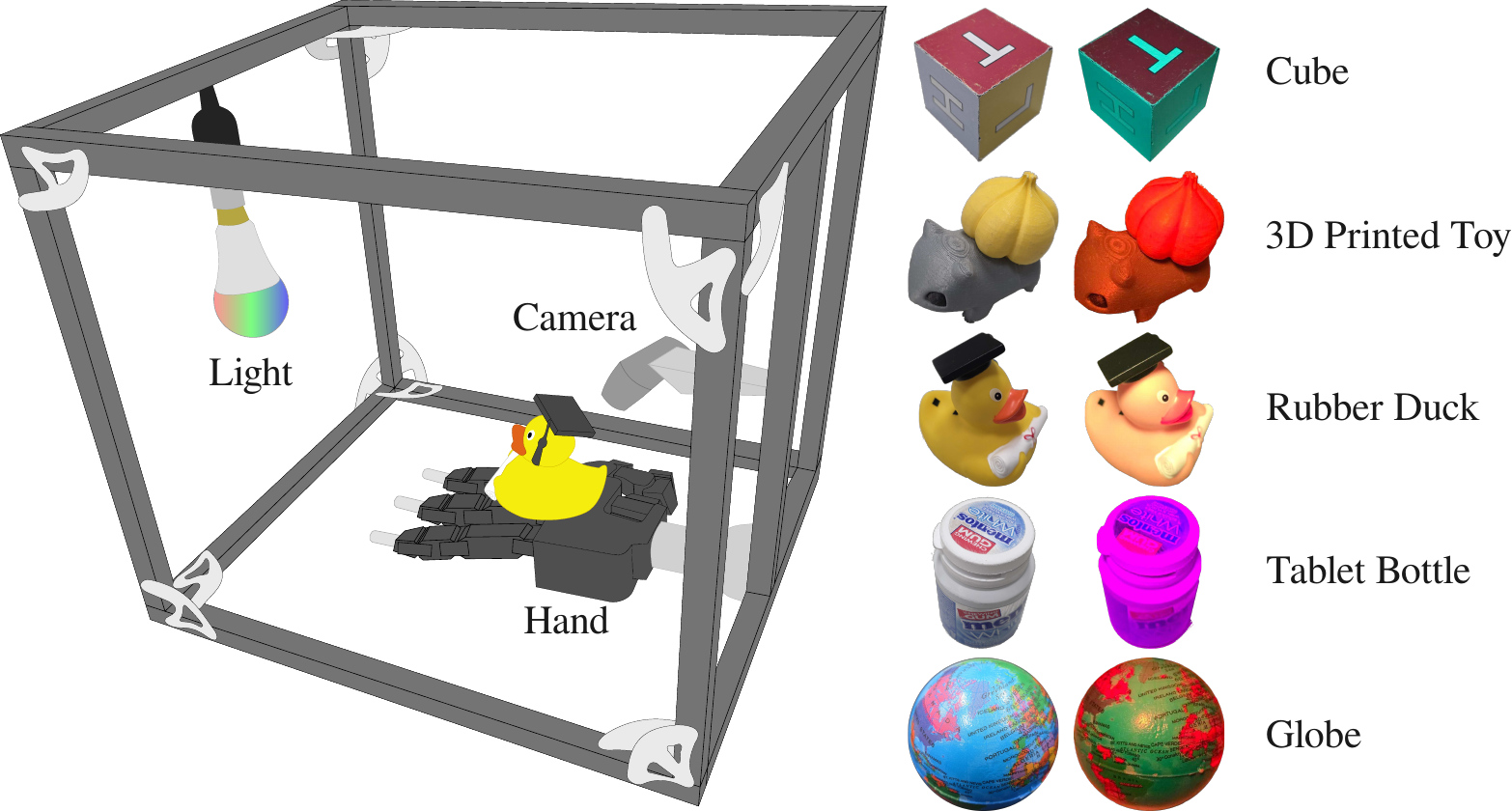}
    \caption{Left: The experimental setup with an RGB camera, an Allegro Hand, and a multi-colored light source for adversarial lighting. Right: The object set displayed under normal lighting (first column) and adversarial lighting (second column).}
    \label{fig:robot_setup}
    \vspace{-16pt}
\end{figure}

We consider a 16-DOF Allegro Hand with a wrist-mounted Intel RealSense D435i camera for visual feedback, as shown in~\cref{fig:robot_setup}. We deploy all models on a single workstation with an Intel Core i9 CPU and NVIDIA RTX 6000 Ada GPU. The policy infers at $30~\si{\hertz}$ and outputs the joint position targets that are tracked by a low-level joint PD controller at $300~\si{\hertz}$. Since the learned pose estimator relies on segmented object inputs, we fine-tune SAM2~\cite{ravi2024sam2} per object to generate precise object masks in real-time.

We evaluate our system on five objects (see~\cref{fig:robot_setup}) exhibiting diverse geometric and physical properties, spanning primitives shapes (\emph{Cube}, \emph{Globe}) to complex, non-convex items (\emph{Tablet Bottle}, \emph{3D Printed Toy}, \emph{Rubber Duck}). For these objects, high-fidelity meshes are obtained using Polycam~\cite{polycam_web}. These are then rendered in NVIDIA Isaac Lab~\cite{mittal2025isaaclab} to generate pose-annotated images for Gaussian Splatting. For each object, we train a pose estimator and control policy in NVIDIA Isaac Lab. Additional training details are provided in the Appendix.

We consider two lighting regimes: nominal (white distant light) and adversarial (low-illumination point sources with dynamic hue shifts). As shown in~\cref{fig:robot_setup}, adversarial conditions present significant visual challenges, including low contrast, specular highlights, and strong color casts, providing a rigorous test of system robustness.

%% file: chapters/4_results.tex
\begin{table*}[t]
\centering
\caption{Evaluation of learned pose estimator on real-world data under nominal and adversarial lighting. Our method is compared against three baselines: \textit{Standard Tiled}, \textit{Randomized Tiled}, and \textit{Naive GS} rendering. We report Average Distance of Model Points (ADD) in mm and a strict accuracy metric ($< 10\text{mm}$ and $< 10^\circ$), averaged over 5 random seeds.} %
\label{tab:pose_estimation_results}
\resizebox{\textwidth}{!}{%

    \begin{tabular}{l|cc|cc|cc|cc|cc|cc}
    \toprule
    \multicolumn{1}{l|}{\textbf{Objects}} & \multicolumn{2}{c|}{\textbf{Cube}} & \multicolumn{2}{c|}{\textbf{3D Printed Toy}} & \multicolumn{2}{c|}{\textbf{Rubber Duck}} & \multicolumn{2}{c|}{\textbf{Tablet Bottle}} & \multicolumn{2}{c|}{\textbf{Globe}} & \multicolumn{2}{c}{\textbf{Mean}} \\
    \midrule
    \textbf{Method} & ADD & Accuracy & ADD & Accuracy & ADD & Accuracy & ADD & Accuracy & ADD & Accuracy & ADD & Accuracy \\
    \midrule
    \multicolumn{13}{c}{\textbf{\textsc{Nominal Conditions}}} \\
    \midrule
    Standard Tiled & $11.9_{\pm 0.86}$ & $57.2_{\pm 8.13}$ & $14.1_{\pm 0.92}$ & $37.7_{\pm 2.84}$ & $9.5_{\pm 0.23}$ & $67.0_{\pm 1.47}$ & $10.0_{\pm 0.65}$ & $45.7_{\pm 4.78}$ & $15.0_{\pm 2.41}$ & $59.1_{\pm 3.68}$ & $12.1_{\pm 1.01}$& $53.3_{\pm 4.18}$ \\
    DR Tiled & $10.5_{\pm 0.75}$ & $69.7_{\pm 9.05}$ & $15.9_{\pm 0.64}$ & $32.5_{\pm 2.78}$ & $9.1_{\pm 0.23}$ & $73.7_{\pm 2.22}$ & $\mathbf{9.0_{\pm 0.52}}$ & $50.7_{\pm 3.56}$ & $16.7_{\pm 1.20}$ & $51.3_{\pm 4.21}$ & $12.2_{\pm 0.67}$& $55.6_{\pm 4.36}$ \\
    Na\"ive GS & $10.4_{\pm 1.35}$ & $59.0_{\pm 4.66}$ & $18.7_{\pm 0.95}$ & $23.3_{\pm 3.18}$ & $11.4_{\pm 0.26}$ & $44.3_{\pm 4.88}$ & $11.2_{\pm 0.63}$ & $48.3_{\pm 4.59}$ & $20.2_{\pm 1.45}$ & $17.2_{\pm 4.61}$ & $14.4_{\pm 0.93}$& $38.4_{\pm 4.38}$ \\
    Ours & $\mathbf{9.1_{\pm 0.51}}$ & $\mathbf{73.1_{\pm 4.73}}$ & $\mathbf{11.3_{\pm 0.82}}$ & $\mathbf{56.5_{\pm 5.77}}$ & $\mathbf{7.9_{\pm 0.24}}$ & $\mathbf{78.0_{\pm 3.97}}$ & $10.2_{\pm 0.84}$ & $\mathbf{51.7_{\pm 5.06}}$ & $\mathbf{12.3_{\pm 0.89}}$ & $\mathbf{67.7_{\pm 3.06}}$ & $\mathbf{10.2_{\pm 0.66}}$& $\mathbf{65.4_{\pm 4.52}}$ \\

    \midrule
    \multicolumn{13}{c}{\textbf{\textsc{Adversarial Conditions}}} \\
    \midrule
    Standard Tiled & $12.6_{\pm 0.74}$ & $56.8_{\pm 4.83}$ & $20.3_{\pm 2.10}$ & $29.1_{\pm 7.05}$ & $15.6_{\pm 0.86}$ & $51.3_{\pm 4.52}$ & $24.1_{\pm 1.58}$ & $25.0_{\pm 3.95}$ & $18.7_{\pm 2.04}$ & $41.9_{\pm 2.68}$ & $18.3_{\pm 1.46}$& $40.8_{\pm 4.61}$ \\
    DR Tiled & $11.5_{\pm 0.89}$ & $57.6_{\pm 4.69}$ & $\mathbf{13.8_{\pm 0.54}}$ & $39.0_{\pm 2.54}$ & $\mathbf{11.5_{\pm 0.61}}$ & $57.7_{\pm 3.89}$ & $14.4_{\pm 1.68}$ & $39.1_{\pm 5.14}$ & $18.6_{\pm 1.07}$ & $42.7_{\pm 4.44}$ & $14.0_{\pm 0.96}$& $47.2_{\pm 4.14}$ \\
    Na\"ive GS & $14.3_{\pm 0.85}$ & $44.3_{\pm 2.30}$ & $21.9_{\pm 0.85}$ & $27.9_{\pm 4.85}$ & $13.9_{\pm 1.25}$ & $57.3_{\pm 4.29}$ & $22.1_{\pm 1.39}$ & $25.0_{\pm 3.35}$ & $21.0_{\pm 1.52}$ & $28.1_{\pm 5.95}$ & $18.6_{\pm 1.17}$& $36.5_{\pm 4.15}$ \\
    Ours & $\mathbf{10.6_{\pm 0.38}}$ & $\mathbf{60.6_{\pm 5.47}}$ & $14.4_{\pm 0.77}$ & $\mathbf{45.9_{\pm 4.58}}$ & $12.2_{\pm 0.53}$ & $\mathbf{62.7_{\pm 3.09}}$ & $\mathbf{13.5_{\pm 0.99}}$ & $\mathbf{46.5_{\pm 3.43}}$ & $\mathbf{13.8_{\pm 0.78}}$ & $\mathbf{65.6_{\pm 2.52}}$ & $\mathbf{12.9_{\pm 0.69}}$& $\mathbf{56.3_{\pm 3.82}}$ \\
    \bottomrule
    \end{tabular}%
    \vspace{-6pt}
}
\end{table*}

\section{Results}

\subsection{Pose Estimation using Different Rendering Pipelines}
\label{sec:pose_results}

To evaluate our pose estimation pipeline, we create a real-world test set for each object using FoundationPose~\cite{Wen2023FoundationPoseU6} as a ground-truth pose labeler. To ensure high-quality labels, we provide FoundationPose with privileged inputs, including object masks, CAD meshes, and high-resolution RGB-D images, and perform multiple refinement iterations. We discard frames where the rendered pose does not visually align with the input image.
Additional details on test dataset is in the Appendix.

We evaluate our proposed pose estimator training pipeline against three baselines, keeping network architecture, training hyperparameters, and dataset sizes constant. The baselines differ solely in their image generation approach:
\begin{enumerate}
    \item \emph{Standard Tiled Rendering:} Isaac Lab's default RTX renderer combined with standard post-process image augmentations.
    \item \emph{Domain Randomized (DR) Tiled Rendering}: Extends the above with randomized scene attributes (background HDRI lighting, material properties such as albedo tint, roughness, and metallic). We use the randomization parameters for material and background from~\cite{singh2024dextrah}.
    \item \emph{Na\"ive Gaussian Splatting:} Our GS-based pipeline without pre-rasterization augmentations, and using only standard post-process image augmentations.
\end{enumerate}

All methods use the same source meshes for geometry-based rendering and Gaussian Splat scene optimization to ensure a fair comparison. We report the Average Distance of Model Points (ADD) and a prediction accuracy metric (error $< 10\text{mm}$ and $< 10^\circ$), averaged over five training seeds.

\paragraph{Nominal Conditions} \cref{tab:pose_estimation_results} details the pose estimation results under nominal lighting. Our method achieves the highest overall performance with a mean accuracy of ${65.4\%}$ and a mean ADD of ${10.2}$~\si{mm}, surpassing standard tiled rendering ($53.3\%$) and the randomized tiled baseline ($55.6\%$). While geometrically simple objects (\eg \textit{Cube}) show similar performance across methods, our approach yields the largest gains on geometrically and texturally complex objects, such as the \textit{3D Printed Toy} ($+24\%$ improvement over randomized tiled rendering) and the \textit{Rubber Duck}.

\paragraph{Adversarial Conditions} Under adversarial lighting, the performance differences between methods become even more pronounced (see~\cref{tab:pose_estimation_results}). Our method demonstrates superior robustness, with a mean accuracy of $\mathbf{56.3\%}$, significantly outperforming the strongest baseline, Randomized Tiled Rendering ($47.2\%$). A critical observation is that the weak performance of the {Na\"ive GS} baseline achieves only $36.5\%$ accuracy. Since this baseline uses the same underlying rendering engine but lacks our specific randomization strategy, this failure highlights that high-fidelity rendering alone is insufficient for generalizing to out-of-distribution visual domains. In contrast, our approach leverages explicit control over scene attributes to generate diverse, challenging training samples. By tailoring these pre-rasterization augmentations to simulate physical lighting variations, which standard 2D image augmentations fail to capture, we maintain performance even in difficult visual scenarios.

\begin{table*}[t]
\centering
\caption{Ablation study of pre-rasterization augmentations under nominal and adversarial conditions for pose estimation. We compare our method with models trained with specific augmentation groups removed. We report Average Distance of Model Points (ADD) in mm and a strict accuracy metric ($< 10\text{mm}$ and $< 10^\circ$), averaged over 5 random seeds.}
\label{tab:ablations_results}
\resizebox{\textwidth}{!}{%
    \begin{tabular}{l|cc|cc|cc|cc|cc|cc}
    \toprule
    \multicolumn{1}{l|}{\textbf{Objects}} & \multicolumn{2}{c|}{\textbf{Cube}} & \multicolumn{2}{c|}{\textbf{3D Printed Toy}} & \multicolumn{2}{c|}{\textbf{Rubber Duck}} & \multicolumn{2}{c|}{\textbf{Tablet Bottle}} & \multicolumn{2}{c|}{\textbf{Globe}} & \multicolumn{2}{c}{\textbf{Mean}} \\
    \midrule
    \textbf{Method} & ADD & Accuracy & ADD & Accuracy & ADD & Accuracy & ADD & Accuracy & ADD & Accuracy & ADD & Accuracy \\
    \midrule
    \multicolumn{13}{c}{\textbf{\textsc{Nominal Conditions}}} \\
    \midrule
    w/o Random Noise & $11.9_{\pm 3.24}$ & $61.5_{\pm 11.21}$ & $13.5_{\pm 1.14}$ & $37.2_{\pm 9.13}$ & $8.2_{\pm 0.39}$ & $\mathbf{80.9_{\pm 2.88}}$ & $10.7_{\pm 0.84}$ & $46.7_{\pm 3.33}$ & $14.5_{\pm 1.17}$ & $66.5_{\pm 5.32}$ & $11.8_{\pm 1.36}$& $58.6_{\pm 6.37}$ \\
    w/o Spatial Clustering & $9.2_{\pm 0.91}$ & $\mathbf{74.9_{\pm 8.49}}$ & $12.9_{\pm 0.86}$ & $44.7_{\pm 7.06}$ & $8.0_{\pm 0.22}$ & $73.5_{\pm 3.27}$ & $16.3_{\pm 1.38}$ & $39.3_{\pm 5.64}$ & $13.8_{\pm 2.15}$ & $51.3_{\pm 5.73}$ & $12.0_{\pm 1.10}$& $56.7_{\pm 6.04}$ \\
    w/o Color Clustering & $9.2_{\pm 0.41}$ & $71.0_{\pm 5.94}$ & $13.6_{\pm 1.60}$ & $49.4_{\pm 2.42}$ & $8.2_{\pm 0.59}$ & $77.4_{\pm 5.64}$ & $20.4_{\pm 0.56}$ & $34.3_{\pm 4.03}$ & $14.5_{\pm 1.69}$ & $64.8_{\pm 3.09}$ & $13.2_{\pm 0.97}$& $59.4_{\pm 4.22}$ \\
    w/o Global Shift & $11.2_{\pm 1.26}$ & $67.2_{\pm 3.85}$ & $\mathbf{10.9_{\pm 1.13}}$ & $55.7_{\pm 6.58}$ & $9.1_{\pm 0.41}$ & $70.9_{\pm 3.85}$ & $39.5_{\pm 2.84}$ & $10.3_{\pm 2.45}$ & $13.9_{\pm 0.25}$ & $52.1_{\pm 5.17}$ & $16.9_{\pm 1.18}$& $51.2_{\pm 4.38}$ \\
    Ours & $\mathbf{9.1_{\pm 0.51}}$ & $73.1_{\pm 4.73}$ & $11.3_{\pm 0.82}$ & $\mathbf{56.5_{\pm 5.77}}$ & $\mathbf{7.9_{\pm 0.24}}$ & $78.0_{\pm 3.97}$ & $\mathbf{10.2_{\pm 0.84}}$ & $\mathbf{51.7_{\pm 5.06}}$ & $\mathbf{12.3_{\pm 0.89}}$ & $\mathbf{67.7_{\pm 3.06}}$ & $\mathbf{10.2_{\pm 0.66}}$& $\mathbf{65.4_{\pm 4.52}}$ \\
    \midrule
    \multicolumn{13}{c}{\textbf{\textsc{Adversarial Conditions}}} \\
    \midrule
    w/o Random Noise & $13.2_{\pm 0.77}$ & $49.7_{\pm 5.46}$ & $15.8_{\pm 1.34}$ & $40.7_{\pm 4.62}$ & $12.9_{\pm 0.39}$ & $52.7_{\pm 4.29}$ & $13.9_{\pm 0.78}$ & $41.5_{\pm 1.10}$ & $16.3_{\pm 0.86}$ & $58.4_{\pm 3.83}$ & $14.4_{\pm 0.83}$& $48.6_{\pm 3.86}$ \\
    w/o Spatial Clustering & $13.5_{\pm 1.40}$ & $46.2_{\pm 7.00}$ & $16.4_{\pm 1.20}$ & $39.3_{\pm 7.13}$ & $14.0_{\pm 0.69}$ & $51.7_{\pm 4.22}$ & $17.6_{\pm 0.84}$ & $35.6_{\pm 3.00}$ & $16.2_{\pm 1.59}$ & $39.6_{\pm 5.92}$ & $15.5_{\pm 1.14}$& $42.5_{\pm 5.45}$ \\
    w/o Color Clustering & $14.4_{\pm 0.68}$ & $47.3_{\pm 6.63}$ & $16.7_{\pm 1.60}$ & $41.2_{\pm 4.03}$ & $14.8_{\pm 0.80}$ & $46.7_{\pm 3.16}$ & $19.9_{\pm 1.15}$ & $29.4_{\pm 4.46}$ & $16.2_{\pm 1.46}$ & $58.9_{\pm 3.93}$ & $16.4_{\pm 1.14}$& $44.7_{\pm 4.44}$ \\
    w/o Global Shift & $24.9_{\pm 1.29}$ & $17.8_{\pm 3.16}$ & $22.0_{\pm 1.54}$ & $25.4_{\pm 2.54}$ & $20.0_{\pm 1.89}$ & $27.7_{\pm 1.70}$ & $30.2_{\pm 1.63}$ & $12.6_{\pm 2.39}$ & $17.5_{\pm 0.52}$ & $34.4_{\pm 2.42}$ & $22.9_{\pm 1.37}$& $23.6_{\pm 2.44}$ \\
    Ours & $\mathbf{10.6_{\pm 0.38}}$ & $\mathbf{60.6_{\pm 5.47}}$ & $\mathbf{14.4_{\pm 0.77}}$ & $\mathbf{45.9_{\pm 4.58}}$ & $\mathbf{12.2_{\pm 0.53}}$ & $\mathbf{62.7_{\pm 3.09}}$ & $\mathbf{13.5_{\pm 0.99}}$ & $\mathbf{46.5_{\pm 3.43}}$ & $\mathbf{13.8_{\pm 0.78}}$ & $\mathbf{65.6_{\pm 2.52}}$ & $\mathbf{12.9_{\pm 0.69}}$& $\mathbf{56.3_{\pm 3.82}}$ \\
    \bottomrule
    \end{tabular}%
}
\end{table*}

\paragraph{Computational Efficiency} Beyond data quality, our integrated 3DGS pipeline provides significant computational advantages over standard rendering solutions. Compared to Isaac Lab's tiled renderer, it achieves a 1.6$\times$ faster rendering throughput on an RTX 6000 Ada. Furthermore, it is substantially more memory-efficient: rendering a batch of 1,024 environments consumes only 12 GB of VRAM, compared to the prohibitive 34 GB required by the tiled renderer. Critically, our proposed pre-rasterization augmentations introduce negligible overhead, executing in less than $2$~\si{\milli\second} per batch, and representing only $\approx 4\%$ of the total frame rendering time.

\subsection{Effect of Pre-Rasterization Augmentations}

\begin{figure*}[h]
    \centering
    \includegraphics[width=0.98\textwidth]{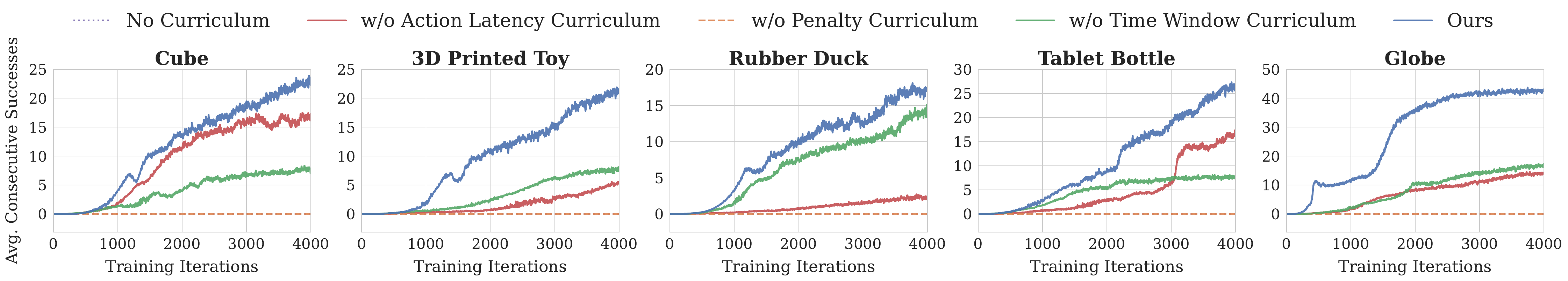}
    \caption{Impact of performance-based curricula on training efficiency. Curves show learning progress for full curriculum compared with ablations with individual components removed. Note that \emph{No Curriculum} and \emph{w/o Penalty Curriculum} coincide at zero.}
    \label{fig:curriculum_results}
    \vspace{-16pt}
\end{figure*}

To isolate the contributions of the pre-rasterization augmentations introduced in~\cref{sec:augmentations}, we perform a ``leave-one-out" ablation study. For each experiment, a pose estimator is trained with data generated with one augmentation group (defined in~\cref{tab:splat_augmentations}) removed and evaluated against the full pipeline. Results on the nominal and adversarial test set are reported in~\cref{tab:ablations_results}.

\paragraph{Nominal Conditions} We observe that the complete augmentation pipeline yields the highest overall robustness. The removal of localized augmentations, specifically \emph{Random Noise} and \emph{Structured Clustering} (Means/SH0), results in moderate performance dips (the accuracy dropping to 57-59\%). This suggests that while fine-grained perturbations refine the decision boundary, they are not the sole drivers of performance under nominal conditions, where lighting is relatively standard.

However, a critical insight emerges from the exclusion of \emph{Global Shift} augmentations. Removing these environmental perturbations causes a significant performance collapse, with mean accuracy dropping to $51.2\%$ and ADD error nearly doubling to $16.9$~\si{mm}. This degradation is particularly acute for the \emph{Tablet Bottle}, which has reflective surfaces. This indicates that even under "nominal" real-world conditions, simulating global variations in exposure, color temperature, and ambient intensity is essential for bridging the sim-to-real gap.

\paragraph{Adversarial Conditions} First, removing \emph{Random Noise} results in a comparatively moderate performance drop (Mean Accuracy: $48.6\%$), suggesting that while unstructured noise aids general robustness, it is not the primary driver of feature learning. In contrast, omitting the structured augmentations leads to significant degradation. The removal of \emph{Spatial Clustering} (3D means) and \emph{Color Clustering} (SH0) drops accuracy to $42.5\%$ and $44.7\%$, respectively, validating that correlated perturbations are essential for modeling localized effects. Most critically, the exclusion of \emph{Global Shift} augmentations causes a catastrophic collapse in performance, with mean accuracy plummeting to just $23.6\%$. This underscores that simulating macro-level environmental changes is the single most important factor for generalizing to diverse real-world lighting conditions.

\subsection{In-hand Reorientation Policy Learning}
\label{sec:rl-policy-results}

\subsubsection{Resource-Efficient Policy Training}
The teacher RL training phase learns in-hand manipulation skills efficiently, scaling across diverse object geometries while using minimal GPU resources. Policies are trained using 24,576 parallel environments.
For primitive geometries (\eg the Cube), the training converges in 26 hours on a single consumer-grade NVIDIA RTX 4090 (24GB VRAM). More complex objects are trained on a dual-GPU setup, requiring 30 GB VRAM and converging in roughly 90 hours due to the increased simulation fidelity of complex contact geometry.
The subsequent student-teacher distillation phase completes in 16 hours for $4,096$ environments on a single NVIDIA RTX 4090 GPU.
Compared to prior work~\cite{handa2023dextreme}, which requires eight A40 GPUs over 60 hours, our pipeline achieves an order-of-magnitude improvement in VRAM efficiency and substantially reduced training time, making high-fidelity RL more accessible for real-world robotics.

\subsubsection{Impact of Performance-based Curriculum}
We evaluate the sample efficiency of our performance-based curriculum by selectively enabling different curriculum components (\cref{sec:curriculum_approach}). \cref{fig:curriculum_results} reports the consecutive successes averaged across all environments during training. Applying all the curriculum components leads to the fastest convergence and highest CS. Removing either the \textit{Action Latency} or \textit{Time Window} components significantly slows learning, particularly for complex geometries such as the \emph{3D Printed Toy} and \emph{Rubber Duck}. Removing the \textit{Regularization Penalty} curriculum causes complete failure, since the policy becomes more conservative and deprioritizes task completion. Similarly, having no curriculum also yields near-zero success. Beyond accelerating convergence, the curriculum also acts as a self-regulating mechanism for reward balancing, eliminating the need for per-object tuning. All teacher RL training and student distillation experiments use the same reward weights and hyperparameters across objects, demonstrating the approach's generality.

\subsection{Hardware Deployment}

\begin{figure}
    \centering
    \includegraphics[width=\linewidth]{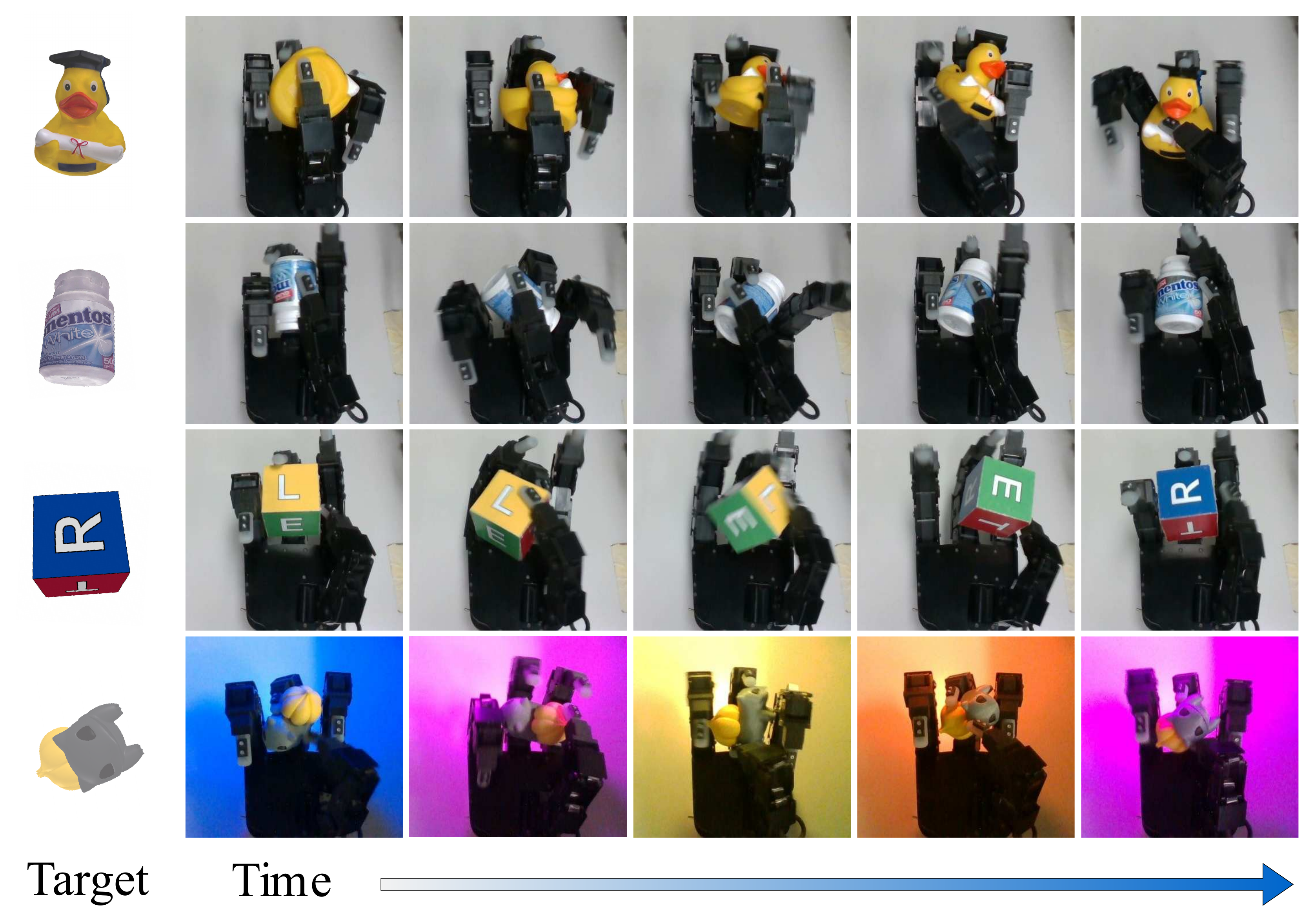}
    \caption{Rollout sequence of the hand reorienting an object to the target pose. The images are from the robot camera feed.}
    \label{fig:rollout}
    \vspace{-14pt}
\end{figure}

Similar to the pose estimation experiments, we validate the sim-to-real deployment of the in-hand reorientation system under nominal and adversarial lighting, as shown in Fig. 6. Consistent with prior work~\cite{andrychowicz2020learning, handa2023dextreme}, we define successful in-hand reorientation when the object orientation error is within $0.4$ radians. We measure \textit{consecutive successes} (CS), \ie, the number of goals reached before a failure occurs (object falls out of the hand). Table IV reports the consecutive successes (CS) averaged over five runs per object and lighting condition.

\paragraph{Comparison to Baselines} Under nominal lighting, our system achieves a mean of $37.6$ consecutive reorientations across objects. It substantially outperforms the only prior vision-based baseline reported on hardware, DeXtreme~\cite{handa2023dextreme}, on the shared \emph{Cube} object ($35.4$ vs.\ $27.8$). %

To further validate the necessity of our tailored perception pipeline, we replaced our estimator with FoundationPose~\cite{Wen2023FoundationPoseU6} during deployment. This configuration resulted in near-total failure, achieving only 0.4 consecutive successes (CS) on average. We attribute this collapse to two factors: (i) FoundationPose operates at approximately 4 Hz, which is too slow for the rapid control loop compared to the $\sim$18 Hz throughput of our estimator, and (ii) frequent tracking loss caused by rapid object motion and severe finger occlusions inherent to in-hand manipulation. This experiment underscores that robust manipulation performance depends not only on policy quality but critically on high-frequency, occlusion-tolerant pose estimation.

\paragraph{Object-Specific Performance} The learned policy demonstrates exceptional robustness on primitive geometries, with the \emph{Globe} exceeding 200 CS in one of the trials. The performance extends to highly non-convex objects, such as the \emph{3D Printed Toy} and \emph{Rubber Duck}, obtaining more than 20 CS on average.  We qualitatively observe smoother manipulation behaviors with compliant objects (e.g., \emph{Globe}, \emph{Duck}), suggesting that inherent material damping aids stability despite not being explicitly modeled in simulation. However, we observe a notable sim-to-real gap for the \emph{Tablet Bottle}. We attribute this degradation to unmodeled friction effects, particularly the extremely low surface friction introduced by its label.

\paragraph{Robustness to Adversarial Lighting} Under severe adversarial lighting conditions, the policy maintained an average of consecutive successes of $\sim$25. To our knowledge, this is the first demonstration of sustained dexterous in-hand manipulation under such extreme visual perturbations. Although performance decreases relative to nominal lighting, the decline highlights the tight coupling between perception and control. The reduction in overall task success disproportionately exceeds the degradation in pose estimation accuracy (\cref{sec:pose_results}). This indicates that even small perceptual errors can compound into significant control failures.

\begin{table}
    \centering
    \caption{Real-world deployment on different objects.}
    \label{tab:real_experiments_results}
    \resizebox{\columnwidth}{!}{%
        \begin{tabular}{l|C{1.8cm}|C{1.8cm}|C{1.8cm}}
        \toprule
        \multicolumn{1}{c|}{} & \multicolumn{3}{|c}{\makecell{\textbf{Consecutive Successes (Over Five Runs)}}} \\
        \midrule
        \multicolumn{1}{l|}{\textbf{Objects}} & \multicolumn{1}{c|}{\makecell{DeXtreme~\cite{handa2023dextreme}}} & \multicolumn{1}{c|}{\makecell{Ours \\ (Nominal)}} & \multicolumn{1}{c}{\makecell{Ours \\ (Adversarial)}} \\
        \midrule
        \textbf{Cube}                       & $27.8_{\pm 19.0}$ & ${35.4_{\pm 13.8}}$ & $25.6_{\pm 8.9}$   \\
        \textbf{3D\ Printed\ Toy}           & -                 & ${28.2_{\pm 12.6}}$ & $12.0_{\pm 6.9}$ \\
        \textbf{Rubber Duck}                & -                 & ${24.2_{\pm 15.3}}$ & $9.0_{\pm 5.0}$  \\
        \textbf{Tablet Bottle}              & -                 & ${12.6_{\pm 8.8}}$ & $4.2_{\pm 0.7}$  \\
        \textbf{Globe}                      & -                 & ${87.6_{\pm 41.4}}$ & $76.2_{\pm 66.2}$  \\
            \midrule
        \textbf{Mean}              & -                 & ${37.6_{\pm 21.8}}$ & $25.4_{\pm 30.1}$   \\
        \bottomrule
        \end{tabular}%
    }
    \vspace{-10pt}
\end{table}

%% file: chapters/5_conclusion.tex
\section{Conclusion} 
\label{sec:conclusion}
In this work, we presented a framework for robust in-hand reorientation using monocular RGB vision. By integrating 3D Gaussian Splatting directly into the simulation loop, we substantially reduced the computational cost of high-fidelity rendering.  Our novel \textit{pre-rasterization augmentations} introduce structured diversity into static Gaussian scenes, effectively bridging the visual sim-to-real gap. This approach enabled the training of a perception module for object pose estimation that is robust to both nominal and severe adversarial lighting conditions. Deploying the resulting system on a multi-fingered hand, we achieved an average of over 25 consecutive successful reorientations across diverse object geometries, even under adversarial lighting.

Our findings underscore a critical insight for the field: the primary bottleneck in real-world dexterity often lies less in control complexity and more in perceptual fidelity. We show that when vision is modeled with sufficient physical grounding, it can support complex and precise manipulation tasks, previously dominated by proprioceptive or tactile-based approaches. Nevertheless, vision is fundamentally constrained by finger occlusions and its inability to directly sense contact forces. A promising direction for future work is the integration of dense visual feedback with high-frequency tactile sensing to better handle unmodeled surface properties. Moreover, while our method excels at instance-specific manipulation, extending this pipeline to support broader generalization is a key step toward truly general-purpose dexterous manipulation.

%% file: chapters/A_appendix.tex
\section{Additional Results}

\subsection{Simulation Results}
To quantify the distillation gap and robustness against observation noise, we evaluate the teacher and student policies across randomized simulation environments parameterized by the domain randomization ranges used during training. Results are summarized in \cref{tab:sim_experiments_results}.

As expected, the teacher policy, with access to ground-truth dynamics via privileged system information, consistently outperforms the student policy across all geometries, even when the student is provided with noiseless observations. This performance delta underscores the value of explicit physical state information (e.g., friction, mass) that cannot be perfectly inferred from proprioception and exteroceptive information alone. Crucially, however, the student policy exhibits minimal performance degradation when transitioning from noiseless to noisy exteroceptive observations. This stability indicates that the distillation process successfully imparts robustness against the noise model, enabling the student to filter observation noise effectively while retaining the behaviour of the teacher policy.

When comparing simulation results to hardware deployments, we observe a quantitative performance gap, a phenomenon common in contact-rich manipulation tasks~\cite{andrychowicz2020learning, yang2024anyrotate}. We attribute this sim-to-real gap to inevitable physical disparities, including inaccurate contact models, actuator dynamics, and material properties. A notable instance of this is the \emph{Tablet Bottle}, where unmodeled low-surface friction significantly impacted real-world controllability despite high simulation success. Additionally, although our perception noise model significantly enhances robustness, it cannot perfectly replicate the full spectrum of stochastic sensor noise found in the real-world sensors, leading to a natural drop in performance on hardware.

\subsection{Belief State Analysis}
\begin{figure}[t]
    \centering
    \includegraphics[width=\columnwidth]{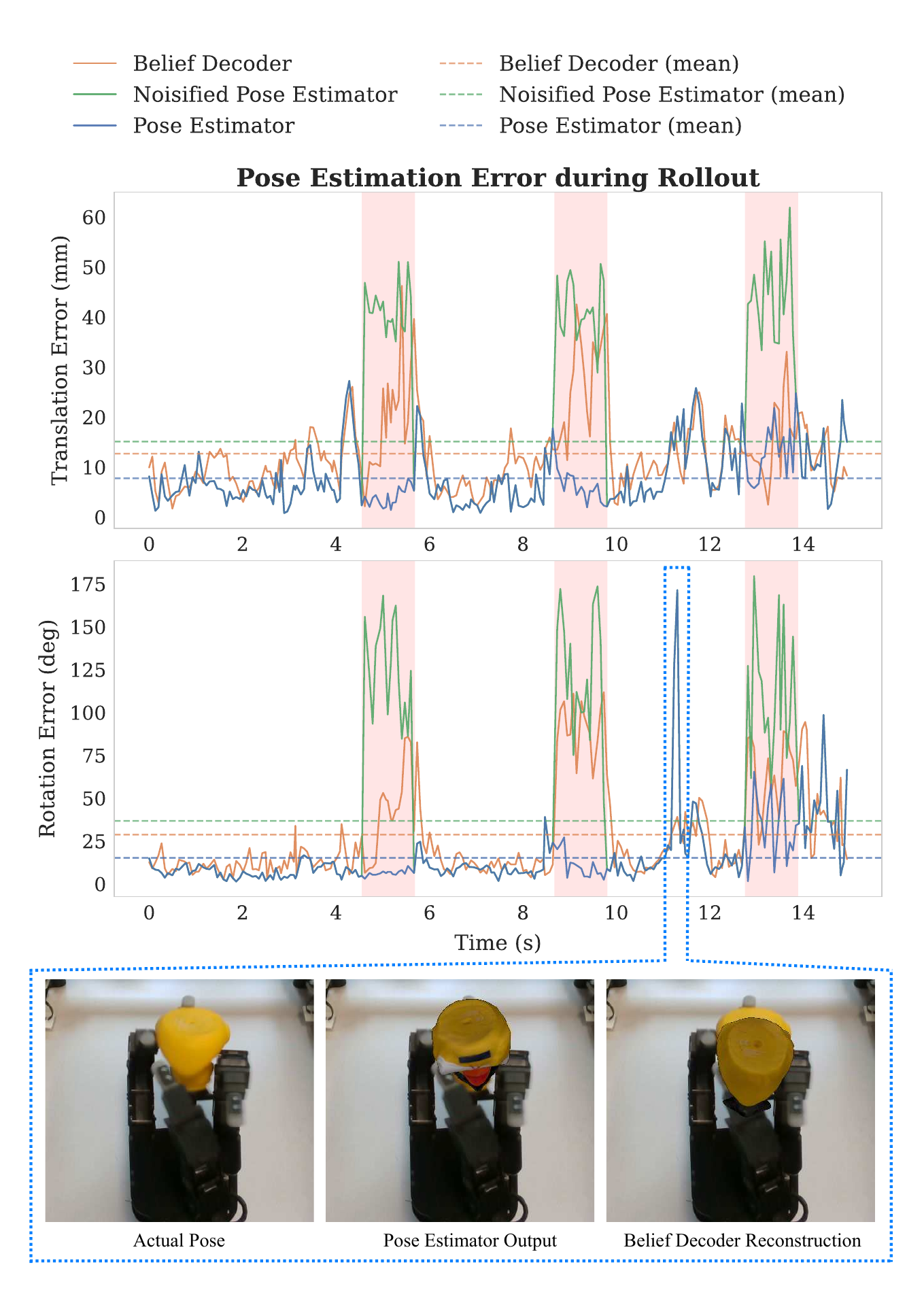}
    \caption{\textbf{Top:} Temporal evolution of translation and rotation errors during a real-world rollout. Red regions indicate intervals where artificial noise is injected into the pose estimator input. The belief decoder (orange) effectively filters these high-frequency perturbations, maintaining significantly lower error compared to the corrupted input (green). \textbf{Bottom:} Visualization of a specific failure case (corresponding to the blue box in the plot). The belief decoder successfully maintains a stable pose estimate close to the ground truth, effectively rejecting a catastrophic 180$^\circ$ flip returned by the pose estimator.}
    \label{fig:belief_decoder_analysis}
    \vspace{-6pt}
\end{figure}

To evaluate the noise rejection capabilities of the belief decoder, we analyze the reconstructed exteroceptive state derived from the belief latent $z_t$. During real-world rollouts, we artificially corrupt the pose estimator's output with synthetic noise at intermittent 1-second intervals. We then compare the error of the belief decoder's reconstruction against the raw pose estimator input, utilizing FoundationPose~\cite{Wen2023FoundationPoseU6} as the offline ground truth reference. The resulting position and rotation errors are visualized in \cref{fig:belief_decoder_analysis}, where red shaded regions indicate phases of noise injection.

\begin{table}[h]
    \centering
    \caption{Simulation results on different objects. We evaluate the policy across 256 environments with randomized physical conditions (DR) and report the mean and standard deviation over five episodes.}
    \label{tab:sim_experiments_results}
    \resizebox{\columnwidth}{!}{%
        \begin{tabular}{l|C{1.8cm}|C{1.8cm}|C{1.8cm}}
        \toprule
        \multicolumn{1}{c|}{} & \multicolumn{3}{|c}{\makecell{\textbf{Consecutive Successes}}} \\
        \midrule
        \multicolumn{1}{l|}{\textbf{Objects}} & \multicolumn{1}{c|}{\makecell{Teacher Policy \\ (w/o obs noise)}} & \multicolumn{1}{c|}{\makecell{Student Policy \\ (w/o obs noise)}} & \multicolumn{1}{c}{\makecell{Student Policy \\ (with obs noise)}} \\
        \midrule
        \textbf{Cube}                       & $111.4_{\pm 24.7}$ & ${92.1_{\pm 5.7}}$ & $82.3_{\pm 8.4}$   \\
        \textbf{3D\ Printed\ Toy}           & $106.0_{\pm 25.0}$ & ${74.9_{\pm 4.7}}$ & $74.6_{\pm 6.2}$ \\
        \textbf{Rubber Duck}                & $97.1_{\pm 14.1}$ & ${46.2_{\pm 1.8}}$ & $41.6_{\pm 2.3}$  \\
        \textbf{Tablet Bottle}              & $118.4_{\pm 2.8}$ & ${77.0_{\pm 4.2}}$ & $69.4_{\pm 2.9}$  \\
        \textbf{Globe}                      & $163.6_{\pm 3.4}$ & ${138.1_{\pm 2.4}}$ & $129.5_{\pm 5.9}$  \\
        \midrule
        \textbf{Mean}                       & $119.3_{\pm 17.05}$ & $85.7_{\pm 4.03}$ & $79.5_{\pm 5.61}$ \\
        \bottomrule
        \end{tabular}%
    }
    \vspace{-10pt}
\end{table}

Under nominal conditions, the raw pose estimator yields lower error than the belief decoder, as indicated by the mean error lines. This is an expected artifact of the decoder's training on high-variance noise, which induces a conservative smoothing bias in the predictions. However, during noise injection phases, the belief decoder demonstrates superior robustness, yielding significantly lower errors than the corrupted sensor input. It effectively filters high-amplitude noise, maintaining stable state estimates for multiple timesteps before drift accumulates. A critical instance of this robustness is highlighted in the region highlighted by the blue box. While the pose estimator suffers a catastrophic 180$^\circ$ flip, the belief decoder successfully rejects this outlier. This confirms that the recurrent architecture effectively functions as a temporal filter, mitigating high-frequency perturbations and preventing tracking divergence during sensor failures.

\subsection{Pose Estimation Results for Rollouts}

\begin{table}[h]
    \centering
    \caption{Pose estimation during real-world hardware deployment. We report the \makebox{translation error (in mm)} and the rotation error (in degrees). We also report Pearson correlation coefficient between the occlusion ratio and errors.}
    \label{tab:rollout_pose_estimation}
    \resizebox{\columnwidth}{!}{%
        \begin{tabular}{l|C{1.8cm}|C{1.8cm}|C{1.8cm}|C{1.8cm}}
        \toprule
        \multicolumn{1}{l|}{\textbf{Objects}} & \multicolumn{1}{c|}{\makecell{Trans Error}} & \multicolumn{1}{c|}{\makecell{Rot Error}} & \multicolumn{1}{c|}{\makecell{Trans Error \\ Correlation}} & \multicolumn{1}{c}{\makecell{Rot Error \\ Correlation}} \\
        \midrule
        \textbf{Cube}                       & $9.05$ & $14.6$ & $0.4$ & $0.38$   \\
        \textbf{3D\ Printed\ Toy}           & $11.14$ & $33.59$ & $0.42$ & $0.12$ \\
        \textbf{Rubber Duck}                & $8.85$ & $18.84$ & $0.02$ & $0.17$  \\
        \textbf{Tablet Bottle}              & $10.9$ & $38.27$ & $0.14$ & $0.08$  \\
        \textbf{Globe}                      & $12.01$ & $32.42$ & $0.04$ & $0.20$  \\
        \midrule
        \textbf{Mean}                       & $10.39$ & $27.54$ & $0.20$ & $0.19$ \\
        \bottomrule
        \end{tabular}%
    }
    \vspace{-10pt}
\end{table}

We quantify the errors of our pose estimator during hardware deployment, with results detailed in \cref{tab:rollout_pose_estimation}. The system demonstrates strong overall performance, achieving a mean translation error of $10.39$ mm and a mean rotation error of $27.54^\circ$. Performance varies across geometries: the \textit{Rubber Duck} yields the lowest translation error ($8.85$ mm), while the \textit{Cube} exhibits the lowest rotational error ($14.6^\circ$). Conversely, objects with complex or symmetric properties, such as the \textit{Tablet Bottle} and \textit{Globe}, show higher rotational variance ($38.27^\circ$ and $32.42^\circ$, respectively).

To evaluate robustness against hand-object interference, we compute the Pearson correlation coefficient between the error metrics and the instantaneous occlusion ratio. The mean correlations are remarkably low ($0.20$ for translation and $0.19$ for rotation), suggesting that pose accuracy is not strongly degraded by partial occlusions. While the \textit{Cube} shows moderate sensitivity (correlation $\approx 0.4$), performance on complex objects like the \textit{Rubber Duck} and \textit{Globe} is effectively decoupled from occlusion levels (translation correlations of $0.02$ and $0.04$, respectively), validating that the pose estimator successfully leverages local features when global geometry is occluded.

\subsection{Pose Estimation Errors}

We report explicit Translation Error (in mm) and Rotation Error (in degrees) for the pose estimator across two studies. In the baseline comparison, reported in \cref{tab:error_pose_estimation_results}, we compare our approach against Standard Tiled, Domain Randomized, and Naive GS rendering. While translation errors remain comparable across methods in nominal conditions, our method demonstrates significantly superior rotational stability. Under adversarial lighting, our pipeline achieves a mean rotation error of 14.6$^\circ$, outperforming other methods by a significant margin. In \cref{tab:error_ablation_results}, we quantify the contribution of each augmentation component. Similar to \cref{sec:pose_results} results identify Global Shift as the most critical factor; its removal causes a catastrophic degradation in adversarial rotation error (spiking to 38.9$^\circ$). Structured perturbations (Spatial and Color Clustering) are shown to be essential for maintaining precision, whereas unstructured Random Noise has a comparatively minor impact on final pose errors.

\begin{table*}[t]
\centering
\caption{Evaluation of learned pose estimator on real-world data under nominal and adversarial lighting. Our method is compared against three baselines: \textit{Standard Tiled}, \textit{Randomized Tiled}, and \textit{Naive GS} rendering. We report the \makebox{translation error (in mm)} and the rotation error (in degrees), averaged over 5 random seeds.}
\label{tab:error_pose_estimation_results}
\resizebox{\textwidth}{!}{%
    \begin{tabular}{l|cc|cc|cc|cc|cc|cc}
    \toprule
    {\textbf{Objects}} & \multicolumn{2}{c|}{\textbf{Cube}} & \multicolumn{2}{c|}{\textbf{3D Printed Toy}} & \multicolumn{2}{c|}{\textbf{Rubber Duck}} & \multicolumn{2}{c|}{\textbf{Tablet Bottle}} & \multicolumn{2}{c|}{\textbf{Globe}} & \multicolumn{2}{c}{\textbf{Mean}} \\
    \midrule
    {\textbf{Method}} & Trans Error & Rot Error & Trans Error & Rot Error & Trans Error & Rot Error & Trans Error & Rot Error & Trans Error & Rot Error & Trans Error & Rot Error \\
    \midrule
    \multicolumn{13}{c}{\textbf{\textsc{Nominal Conditions}}} \\
    \midrule
    Standard Tiled & $10.8_{\pm 0.78}$ & $5.6_{\pm 1.57}$ & $11.2_{\pm 0.32}$ & $19.2_{\pm 3.79}$ & $8.1_{\pm 0.22}$ & $7.5_{\pm 0.75}$ & $8.2_{\pm 0.39}$ & $13.3_{\pm 1.41}$ & $13.9_{\pm 2.51}$ & $8.4_{\pm 0.59}$ & $10.4_{\pm 1.20}$ & $10.8_{\pm 1.99}$ \\
    Domain Randomized & $10.0_{\pm 0.80}$ & $\mathbf{4.0_{\pm 0.15}}$ & $15.4_{\pm 0.74}$ & $\mathbf{8.8_{\pm 0.78}}$ & $8.3_{\pm 0.26}$ & $5.0_{\pm 0.20}$ & $7.8_{\pm 0.13}$ & $\mathbf{10.2_{\pm 1.29}}$ & $14.5_{\pm 1.29}$ & $14.5_{\pm 2.31}$ & $11.2_{\pm 0.77}$ & $8.5_{\pm 1.24}$ \\
    Naive GS & $8.7_{\pm 1.25}$ & $6.9_{\pm 1.22}$ & $15.3_{\pm 0.96}$ & $23.2_{\pm 2.04}$ & $10.6_{\pm 0.32}$ & $6.0_{\pm 0.27}$ & $8.6_{\pm 0.54}$ & $16.0_{\pm 3.54}$ & $18.8_{\pm 1.50}$ & $11.7_{\pm 1.38}$ & $12.4_{\pm 1.01}$ & $12.8_{\pm 2.01}$ \\
    Ours & $\mathbf{8.2_{\pm 0.64}}$ & $4.8_{\pm 0.18}$ & $\mathbf{10.2_{\pm 1.01}}$ & $\mathbf{8.8_{\pm 0.82}}$ & $\mathbf{7.5_{\pm 0.26}}$ & $\mathbf{4.2_{\pm 0.08}}$ & $\mathbf{7.5_{\pm 0.74}}$ & $10.3_{\pm 1.93}$ & $\mathbf{11.7_{\pm 0.83}}$ & $\mathbf{5.3_{\pm 0.55}}$ & $\mathbf{9.0_{\pm 0.74}}$ & $\mathbf{6.7_{\pm 0.97}}$ \\
    \midrule
    \multicolumn{13}{c}{\textbf{\textsc{Adversarial Conditions}}} \\
    \midrule
    Standard Tiled & $\mathbf{9.0_{\pm 0.95}}$ & $12.0_{\pm 1.74}$ & $15.9_{\pm 1.49}$ & $33.4_{\pm 5.14}$ & $9.1_{\pm 0.53}$ & $25.1_{\pm 1.83}$ & $12.4_{\pm 1.03}$ & $57.2_{\pm 4.08}$ & $15.3_{\pm 2.38}$ & $20.7_{\pm 1.39}$ & $12.3_{\pm 1.42}$ & $29.7_{\pm 3.21}$ \\
    Domain Randomized & $9.1_{\pm 0.64}$ & $9.2_{\pm 1.63}$ & $\mathbf{12.2_{\pm 0.58}}$ & $\mathbf{13.3_{\pm 1.79}}$ & $9.1_{\pm 0.38}$ & $\mathbf{10.7_{\pm 1.31}}$ & $8.7_{\pm 0.72}$ & $28.8_{\pm 4.55}$ & $14.4_{\pm 1.37}$ & $26.5_{\pm 2.03}$ & $10.7_{\pm 0.81}$ & $17.7_{\pm 2.55}$ \\
    Naive GS & $10.9_{\pm 0.79}$ & $13.5_{\pm 1.21}$ & $17.4_{\pm 0.66}$ & $37.6_{\pm 2.73}$ & $\mathbf{8.3_{\pm 0.43}}$ & $22.3_{\pm 4.35}$ & $9.5_{\pm 0.76}$ & $56.3_{\pm 4.37}$ & $15.9_{\pm 1.84}$ & $31.1_{\pm 1.80}$ & $12.4_{\pm 1.02}$ & $32.2_{\pm 3.17}$ \\
    Ours & $9.1_{\pm 0.46}$ & $\mathbf{6.6_{\pm 0.78}}$ & $12.7_{\pm 0.61}$ & $17.2_{\pm 1.67}$ & $9.5_{\pm 0.32}$ & $12.0_{\pm 2.07}$ & $\mathbf{7.7_{\pm 0.35}}$ & $\mathbf{26.7_{\pm 2.66}}$ & $\mathbf{12.3_{\pm 0.71}}$ & $\mathbf{10.7_{\pm 2.20}}$ & $\mathbf{10.3_{\pm 0.51}}$ & $\mathbf{14.6_{\pm 1.98}}$ \\
    \bottomrule
    \end{tabular}%
}
\end{table*}

\begin{table*}[t]
\centering
\caption{Ablation study of pre-rasterization augmentations under nominal and adversarial conditions for pose estimation. We compare our method with models trained with specific augmentation groups removed. We report the \makebox{translation error (in mm)} and the rotation error (in degrees), averaged over 5 random seeds.}
\label{tab:error_ablation_results}
\resizebox{\textwidth}{!}{%
    \begin{tabular}{l|cc|cc|cc|cc|cc|cc}
    \toprule
    {\textbf{Objects}} & \multicolumn{2}{c|}{\textbf{Cube}} & \multicolumn{2}{c|}{\textbf{3D Printed Toy}} & \multicolumn{2}{c|}{\textbf{Rubber Duck}} & \multicolumn{2}{c|}{\textbf{Tablet Bottle}} & \multicolumn{2}{c|}{\textbf{Globe}} & \multicolumn{2}{c}{\textbf{Mean}} \\
    \midrule
    {\textbf{Method}} & Trans Error & Rot Error & Trans Error & Rot Error & Trans Error & Rot Error & Trans Error & Rot Error & Trans Error & Rot Error & Trans Error & Rot Error \\
    \midrule
    \multicolumn{13}{c}{\textbf{\textsc{Nominal Conditions}}} \\
    \midrule
    w/o Random Noise & $10.8_{\pm 3.05}$ & $6.8_{\pm 2.27}$ & $12.5_{\pm 0.91}$ & $10.0_{\pm 2.63}$ & $7.7_{\pm 0.44}$ & $4.7_{\pm 0.29}$ & $8.3_{\pm 0.62}$ & $12.6_{\pm 2.20}$ & $14.0_{\pm 1.16}$ & $5.5_{\pm 0.59}$ & $10.7_{\pm 1.55}$ & $7.9_{\pm 1.86}$ \\
    w/o Spatial Clustering & $\mathbf{8.1_{\pm 1.23}}$ & $5.6_{\pm 3.46}$ & $11.7_{\pm 0.93}$ & $10.9_{\pm 1.48}$ & $\mathbf{7.2_{\pm 0.28}}$ & $5.8_{\pm 0.24}$ & $10.7_{\pm 0.71}$ & $26.7_{\pm 3.23}$ & $13.1_{\pm 2.36}$ & $6.6_{\pm 0.86}$ & $10.2_{\pm 1.31}$ & $11.1_{\pm 2.25}$ \\
    w/o Color Clustering & $\mathbf{8.1_{\pm 0.62}}$ & $5.4_{\pm 0.54}$ & $12.2_{\pm 1.78}$ & $11.7_{\pm 1.61}$ & $7.6_{\pm 0.65}$ & $4.7_{\pm 0.16}$ & $14.0_{\pm 0.71}$ & $33.8_{\pm 2.11}$ & $13.7_{\pm 1.81}$ & $6.5_{\pm 0.98}$ & $11.1_{\pm 1.25}$ & $12.4_{\pm 1.29}$ \\
    w/o Global Shift & $8.3_{\pm 0.60}$ & $11.0_{\pm 4.64}$ & $\mathbf{9.9_{\pm 1.10}}$ & $9.5_{\pm 1.69}$ & $8.1_{\pm 0.34}$ & $6.6_{\pm 0.53}$ & $29.5_{\pm 3.50}$ & $67.9_{\pm 1.83}$ & $11.9_{\pm 0.43}$ & $12.4_{\pm 1.10}$ & $13.5_{\pm 1.68}$ & $21.5_{\pm 2.42}$ \\
    Ours & $8.2_{\pm 0.64}$ & $\mathbf{4.8_{\pm 0.18}}$ & $10.2_{\pm 1.01}$ & $\mathbf{8.8_{\pm 0.82}}$ & $7.5_{\pm 0.26}$ & $\mathbf{4.2_{\pm 0.08}}$ & $\mathbf{7.5_{\pm 0.74}}$ & $\mathbf{10.3_{\pm 1.93}}$ & $\mathbf{11.7_{\pm 0.83}}$ & $\mathbf{5.3_{\pm 0.55}}$ & $\mathbf{9.0_{\pm 0.74}}$ & $\mathbf{6.7_{\pm 0.97}}$ \\
    \midrule
    \multicolumn{13}{c}{\textbf{\textsc{Adversarial Conditions}}} \\
    \midrule
    w/o Random Noise & $11.1_{\pm 0.73}$ & $9.3_{\pm 1.68}$ & $14.2_{\pm 1.24}$ & $\mathbf{16.8_{\pm 1.90}}$ & $10.4_{\pm 0.45}$ & $12.2_{\pm 1.18}$ & $8.7_{\pm 0.60}$ & $\mathbf{26.1_{\pm 1.83}}$ & $15.0_{\pm 0.85}$ & $11.5_{\pm 1.15}$ & $11.9_{\pm 0.82}$ & $15.2_{\pm 1.58}$ \\
    w/o Spatial Clustering & $11.2_{\pm 1.34}$ & $9.7_{\pm 1.11}$ & $13.5_{\pm 1.39}$ & $23.6_{\pm 1.15}$ & $10.0_{\pm 0.30}$ & $17.3_{\pm 2.21}$ & $8.9_{\pm 0.72}$ & $39.8_{\pm 2.90}$ & $13.7_{\pm 1.85}$ & $17.1_{\pm 1.51}$ & $11.5_{\pm 1.25}$ & $21.5_{\pm 1.90}$ \\
    w/o Color Clustering & $10.9_{\pm 0.84}$ & $13.2_{\pm 2.58}$ & $14.4_{\pm 1.45}$ & $22.3_{\pm 2.46}$ & $10.8_{\pm 0.44}$ & $17.9_{\pm 2.85}$ & $11.2_{\pm 0.71}$ & $41.8_{\pm 4.49}$ & $14.0_{\pm 1.85}$ & $13.8_{\pm 3.40}$ & $12.3_{\pm 1.18}$ & $21.8_{\pm 3.24}$ \\
    w/o Global Shift & $16.0_{\pm 0.93}$ & $35.4_{\pm 2.40}$ & $16.6_{\pm 1.14}$ & $41.2_{\pm 4.81}$ & $14.7_{\pm 1.71}$ & $25.2_{\pm 3.11}$ & $10.3_{\pm 0.60}$ & $71.7_{\pm 6.55}$ & $13.9_{\pm 0.54}$ & $21.2_{\pm 0.66}$ & $14.3_{\pm 1.07}$ & $38.9_{\pm 4.05}$ \\
    Ours & $\mathbf{9.1_{\pm 0.46}}$ & $\mathbf{6.6_{\pm 0.78}}$ & $\mathbf{12.7_{\pm 0.61}}$ & $17.2_{\pm 1.67}$ & $\mathbf{9.5_{\pm 0.32}}$ & $\mathbf{12.0_{\pm 2.07}}$ & $\mathbf{7.7_{\pm 0.35}}$ & $26.7_{\pm 2.66}$ & $\mathbf{12.3_{\pm 0.71}}$ & $\mathbf{10.7_{\pm 2.20}}$ & $\mathbf{10.3_{\pm 0.51}}$ & $\mathbf{14.6_{\pm 1.98}}$ \\
    \bottomrule
    \end{tabular}%
}
\end{table*}

\section{Method Details}

\subsection{Teacher Training using Reinforcement Learning}

\subsubsection{MDP Formulation}

\paragraph{Action Space} The action space $\mathcal{A} \subseteq \mathbb{R}^{16}$ consists of target joint positions for the 16 independently actuated joints of the Allegro Hand. The policy outputs actions $a_t$, which are scaled to the robot's joint limits and processed via an Exponential Moving Average (EMA) filter $\bar{a}_t = (1 - \alpha) \bar{a}_{t-1} + \alpha a_t$ to ensure smooth motion. The smoothing parameter $\alpha$ of the EMA filter is randomized between (0.08, 0.2) during training to enhance robustness against unmodeled gaps in the robot's actuator dynamics.

\paragraph{Reward Modeling} To guide the policy, we employ a dense reward function $r_t = r_{\text{task}} + r_{\text{reg}}$. The task term $r_{\text{task}}$ incentivizes minimizing orientation error, augmented by a sparse success bonus. To encourage smooth and stable behavior, the regularization term $r_{\text{reg}}$ penalizes aggressive control inputs, high joint velocities, energy consumption and object instability. The various reward terms used for training the teacher policy are listed in ~\cref{tab:rewards}

\begin{table}[t]
    \centering
    \caption{Reward Terms}
    \label{tab:rewards}
    \resizebox{\columnwidth}{!}{
    \begin{tabular}{l r l p{4.9cm}}
        \toprule
        \textbf{Term} & \textbf{Weight} & \textbf{Equation} & \textbf{Description} \\
        \midrule

        \multicolumn{4}{l}{\textbf{\textsc{Task Rewards}}} \\
        \addlinespace[0.3em]
        Orientation Tracking & 1.0   & $(d(\theta) + \epsilon)^{-1}$ 
        & $d(\theta)$: orientation error, $\epsilon=0.1$. \\
        Success Bonus        & 250.0 & $\mathbb{I}(d(\theta) \le \epsilon_{\text{success}})$ 
        & Bonus when $d(\theta) \le \epsilon_{\text{success}}=0.1$\,rad. \\
        \addlinespace[0.6em]

        \multicolumn{4}{l}{\textbf{\textsc{Termination Penalties}}} \\
        \addlinespace[0.3em]
        Object Dropped          & $-10.0$ & $\mathbb{I}(\text{drop})$ 
        & Penalty when the object falls. \\
        \addlinespace[0.6em]

        \multicolumn{4}{l}{\textbf{\textsc{Curriculum Penalties}}} \textit{(Weights are gradually increased during training)} \\
        \addlinespace[0.3em]
        Object Distance & -20.0 & $-\|p_{\text{robot}} - p_{\text{obj}}\|_2$ 
        & Object-robot distance. \\
        Object Velocity      & -1e-3 & $-\|v_{\text{obj}}\|_2$ 
        & $L_2$ norm of object linear velocity. \\
        Joint Velocity       & -8e-2 & $-\|\dot{q}\|_2$ 
        & $L_2$ norm of joint velocities. \\
        Action Magnitude     & -0.80 & $-\|a_t\|_2$ 
        & $L_2$ norm of the action vector. \\
        Action Rate          & -0.12 & $-\|a_t - a_{t-1}\|_2$ 
        & $L_2$ norm of diff. b/w consecutive actions. \\
        Joint Work           & -0.12 & $-\sum |\tau \cdot \dot{q}|$ 
        & Mechanical work. \\
        Joint Torques        & -50.0 & $-\|\tau\|_2$ 
        & $L_2$ norm of applied joint torques. \\

        \bottomrule
    \end{tabular}
    }
\end{table}

\paragraph{Early Terminations} An episode terminates if the object falls from the hand, the agent fails to achieve a success within a 10-second window, or the policy successfully completes a sequence of 50 consecutive reorientations.

\paragraph{Observations} \cref{tab:observations_table} lists out the observation terms in the different observation groups introduced in ~\cref{sec:observation_groups}.

\begin{table}[t]
    \centering
    \caption{Observation Space}
    \label{tab:observations_table}
    \resizebox{\columnwidth}{!}{
    \begin{tabular}{l r p{6.5cm}}
        \toprule
        \textbf{Term} & \textbf{Dim.} & \textbf{Description} \\
        \midrule

        \multicolumn{3}{l}{\textbf{\textsc{Proprioceptive Observations}} $\boldsymbol{\mathcal{O}_{\text{prop}}}$} \\
        \addlinespace[0.3em]
        Joint Positions        & 16 & Measured robot joint angles. \\
        Action History         & 64 & Joint position commands from the last four time steps. \\
        Goal Orientation       & 4  & Target orientation (quaternion). \\
        Palm Link Position     & 3  & Position of the palm link (constant) \\
        Remaining Time         & 1  & Normalized time remaining in the episode. \\
        \addlinespace[0.3em]
        \textit{Group Total}   & \textit{88} & \\
        \addlinespace[0.6em]
        
        \multicolumn{3}{l}{\textbf{\textsc{Exteroceptive Observations}} $\boldsymbol{\mathcal{O}_{\text{extero}}}$} \\
        \addlinespace[0.3em]
        Object Pose            & 7  & Ground-truth object position and orientation (quaternion). \\
        Goal Quaternion Diff.  & 4  & Quaternion representing rotation from object to goal. \\
        \addlinespace[0.3em]
        \textit{Group Total}   & \textit{11} & \\
        \addlinespace[0.6em]

        \multicolumn{3}{l}{\textbf{\textsc{Privileged Observations}} $\boldsymbol{\mathcal{O}_{\text{priv}}}$} \\
        \addlinespace[0.3em]
        Joint Velocities       & 16 & Angular velocities of robot joints. \\
        Joint Torques          & 16 & Actuator-applied joint torques. \\
        Fingertip Forces       & 12 & Net contact forces at the four fingertips. \\
        Object Velocities      & 6  & Linear and angular velocities of the object. \\
        Physical Properties    & 7  & Randomized object/robot scale and object mass. \\
        Scene Gravity          & 3  & Gravity vector. \\
        Actuator Gains         & 32 & Randomized joint stiffness and damping. \\
        Action Properties      & 2  & Randomized EMA parameter $\alpha$ and delay \\
        Random Forces          & 6  & External forces and torques applied to the object. \\
        \addlinespace[0.3em]
        \textit{Group Total}   & \textit{100} & \\
        \addlinespace[0.6em]

        \midrule
        \textbf{Total Observation Dim.} & \textbf{199} & \\
        \bottomrule
    \end{tabular}
    }
\end{table}

\subsubsection{Domain Randomization}
We employ domain randomization across multiple aspects of the simulation to improve robustness against varying physical conditions and facilitate sim-to-real transfer. Physical properties of both the robot and the object, including link and object mass, friction coefficients, and restitution, are randomized to account for inaccuracies in contact dynamics. Geometric properties of the object are varied to capture shape and size variations. We perform system identification for the Allegro hand using ~\cite{bjelonic2025towards} to obtain simulation-accurate values for actuation-related parameters, such as joint stiffness and damping. At each episode, these values are randomized around the identified ranges to account for model mismatch between the simulated and actual robot dynamics. Finally, external disturbances are introduced by applying random forces and perturbing the gravity vector at fixed intervals, to add robustness against unmodeled interactions. Various randomizations and their ranges are listed in Table~XI.

\begin{table}[t]
    \caption{Domain Randomization Parameters}
    \label{tab:randomization}
    \centering
    \resizebox{\columnwidth}{!}{
    \begin{tabular}{p{2.8cm} l l p{3.3cm}}
        \toprule
        \textbf{Parameter} & \textbf{Type} & \textbf{Distribution} & \textbf{Range / Details} \\
        \midrule

        \multicolumn{4}{l}{\textbf{\textsc{Startup Randomization}}} \\
        \addlinespace[0.3em]
        Robot Link Mass      & Scaling   & Log-Uniform & $\times [0.75, 1.5]$ \\
        Robot Link Friction  & Absolute  & Uniform     & Static/Dynamic: $[0.0, 0.3]$ \\
        Fingertip Friction   & Absolute  & Uniform     & Static/Dynamic: $[0.3, 0.8]$ \\
        Robot Restitution    & Absolute  & Uniform     & $[0.0, 0.4]$ \\
        Object Scale         & Scaling   & Uniform     & $\times [0.8, 1.2]$ (per axis) \\
        Object Mass          & Scaling   & Uniform     & $\times [0.5, 1.5]$ \\
        Object Friction      & Absolute  & Uniform     & Static/Dynamic: $[0.3, 0.8]$ \\
        Object Restitution   & Absolute  & Uniform     & $[0.0, 0.4]$ \\
        \addlinespace[0.6em]

        \multicolumn{4}{l}{\textbf{\textsc{Reset Randomization}}} \\
        \addlinespace[0.3em]
        Joint Stiffness      & Scaling   & Log-Uniform & $\times [0.75, 1.5]$ \\
        Joint Damping        & Scaling   & Log-Uniform & $\times [0.75, 1.5]$ \\
        Joint Friction       & Scaling   & Log-Uniform & $\times [0.75, 1.5]$ \\
        Joint Armature       & Scaling   & Log-Uniform & $\times [0.75, 1.5]$ \\
        Joint Limits         & Scaling   & Log-Uniform & Lower/Upper: $\times [0.95, 1.05]$ \\
        \addlinespace[0.6em]

        \multicolumn{4}{l}{\textbf{\textsc{Interval Randomization}}} \\
        \addlinespace[0.3em]
        External Forces      & Additive  & Impulse     & Magnitude: $2.0\times$, Prob.: $0.1$ \\
        Gravity Vector       & Additive  & Uniform     & $\pm 0.5$\,m/s$^2$ every $0$--$15$\,s \\
        \bottomrule
    \end{tabular}
    }
\end{table}

\subsubsection{Policy Architecture and Optimization}

We employ a modified asymmetric actor-critic framework to learn the teacher policy. Unlike standard implementations, where only the critic has access to privileged state information, we provide privileged observations~$\mathcal{O}_{\text{priv}}$ to both the actor and the critic. The asymmetry is instead introduced in the actor's proprioceptive input~$\mathcal{O}_{\text{prop}}^{\text{noisy}}$, which is augmented with noise and latency during training, to induce robustness against sensor inaccuracies encountered in real-world observations.

\begin{table}[t]
    \centering
    \caption{Teacher Policy Architecture and PPO Training Configuration}
    \label{tab:teacher_ppo_config}
    \resizebox{\columnwidth}{!}{%
    \begin{tabular}{l c}
        \toprule
        \textbf{Parameter} & \textbf{Value} \\
        \midrule
        \multicolumn{2}{l}{\textbf{\textsc{Policy Architecture}}} \\
        \addlinespace[0.3em]
        Actor Hidden Layers & $[1024, 1024, 1024, 512]$ \\
        Critic Hidden Layers & $[1024, 1024, 1024, 512]$ \\
        Exte. Encoder Hidden Layers & $[64, 64]$ \\
        Exte. Encoder Latent Dimension & $24$ \\
        Priv. Encoder Hidden Layers & $[256, 256]$ \\
        Priv. Encoder Latent Dimension & $128$ \\
        Activation Function & ELU \\
        \addlinespace[0.6em]
        \multicolumn{2}{l}{\textbf{\textsc{PPO Training Parameters}}} \\
        \addlinespace[0.3em]
        Steps per Environment & $24$ \\
        Discount Factor $\gamma$ & $0.998$ \\
        GAE Parameter $\lambda$ & $0.95$ \\
        Learning Rate & $1 \times 10^{-3}$ \\
        Learning Rate Schedule & Adaptive \\
        Clip Range & $0.2$ \\
        Value Loss Coefficient & $0.5$ \\
        Entropy Coefficient & $0.002$ \\
        Learning Epochs per Iteration & $5$ \\
        Mini-batches & $12$ \\
        Target KL Divergence & $0.01$ \\
        \addlinespace[0.6em]
        \bottomrule
    \end{tabular}
    }
\end{table}

To effectively fuse heterogeneous modalities, we utilize a multi-encoder network architecture. Exteroceptive and privileged observation groups are processed by dedicated Multi-Layer Perceptrons (MLPs) to extract latent embeddings $z_{\text{exte}} \in \mathbb{R}^{24}$ and $z_{\text{priv}} \in \mathbb{R}^{128}$, with encoder hidden dimensions of $[64, 64]$ and $[256, 256]$, respectively. These embeddings are concatenated with normalized proprioceptive observations and passed to the primary backbone. We observed a strong correlation between network capacity and task performance; consequently, both the actor and critic are parameterized as deep MLPs with hidden units $[1024, 1024, 1024, 512]$ and ELU activations. Policy optimization is performed using the Proximal Policy Optimization (PPO) algorithm~\cite{Schulman2017ProximalPO}, implemented within the RSL-RL library~\cite{schwarke2025rslrl}. The policy architecture and training hyperparameters are listed in Table XII.

\subsection{Student Training using Distillation}

\subsubsection{Student Policy} 
The parameters for the student policy architecture and training are listed in Table XIII.

\begin{table}[t]
    \caption{Student Policy and Distillation Hyperparameters}
    \label{tab:distillation_table}
    \centering
    \resizebox{\columnwidth}{!}{%
    \begin{tabular}{l c}
        \toprule
        \textbf{Parameter} & \textbf{Value} \\
        \midrule

        \multicolumn{2}{l}{\textbf{\textsc{Student Architecture}}} \\
        \addlinespace[0.3em]
        Actor MLP                    & $[1024, 1024, 512, 512]$ \\
        Exteroceptive MLP            & $[256, 256]$ \\
        Exteroceptive Latent Dim     & $64$ \\
        Privileged Latent Dim        & $256$ \\
        Activation Function          & ELU \\
        Initial Action Noise Std.    & $0.02$ \\
        \addlinespace[0.6em]

        \multicolumn{2}{l}{\textbf{\textsc{Belief Encoder and Decoder}}} \\
        \addlinespace[0.3em]
        RNN Hidden Dimension         & $256$ \\
        Number of RNN Layers         & $2$ \\
        Latent Hidden Dimensions     & $[256, 256]$ \\
        Attention Gate Dimensions    & $[128, 128]$ \\
        Exteroceptive Decoder MLP    & $[256, 256]$ \\
        Privileged Decoder MLP       & $[256, 256]$ \\
        \addlinespace[0.6em]

        \multicolumn{2}{l}{\textbf{\textsc{Distillation Algorithm}}} \\
        \addlinespace[0.3em]
        Optimizer                    & AdamW \\
        Learning Rate                & $3.0 \times 10^{-4}$ \\
        Number of Learning Epochs    & $32$ \\
        Mini-batches                 & $1$ \\
        Backpropagation Length       & $45$ steps \\
        DAgger Mixing Ratio          & $0.9$ \\
        Mixing Ratio Decay           & $0.95$ \\
        Reconstruction Loss Coefficient & $0.2$ \\
        Decoder L1 Loss Coefficient & $0.2$ \\
        Decoder Exteroceptive Loss Coefficient & $2.0$ \\
        \bottomrule
    \end{tabular}
    }
\end{table}

\subsubsection{Online DAgger}
We employ an online variant of DAgger~\cite{ross2011reduction} to mitigate the covariate shift between states induced by the student's and teacher's actions. During data collection, we generate trajectories by stochastically mixing the teacher's and student's actions. 
At each timestep, the action applied to the robot is selected from the teacher's policy with \mbox{probability~$\beta$~(mixing ratio)}, and from the student's policy with probability~$1-\beta$. Regardless of which action is executed, the student is trained to predict the teacher's optimal action for the visited state. To stabilize the early phases of training, we initialize $\beta$ to a high value ($0.9$) and exponentially decay it after each iteration, gradually transitioning full control to the student as its performance improves.

\begin{table}[t]
    \centering
    \caption{Perception Noise Generator Parameters}
    \label{tab:noise}
    \resizebox{\columnwidth}{!}{%
    \begin{tabular}{l l p{3.6cm}}
        \toprule
        \textbf{Noise Component} & \textbf{Distribution / Type} & \textbf{Parameters / Range} \\
        \midrule

        \multicolumn{3}{l}{\textbf{\textsc{Object Position}}} \\
        \addlinespace[0.3em]
        Temporal Downsampling & Discrete Sampling ($k$) 
        & Update period: $1$--$3$ steps \\
        Stochastic Jitter     & Bernoulli Delay ($p$) 
        & Delay probability: $0.0$--$0.1$ \\
        Tracking Failure     & Random Replacement 
        & Failure probability: $0.0$--$0.3$ \\
        Biased Noise          & Additive Uniform 
        & Noise: $\mathcal{U}[-12, 12]$\,mm \newline
          Bias:  $\mathcal{U}[-12, 12]$\,mm \\
        \addlinespace[0.6em]

        \multicolumn{3}{l}{\textbf{\textsc{Object Orientation}}} \\
        \addlinespace[0.3em]
        Temporal Downsampling & Discrete Sampling ($k$) 
        & Update period: $1$--$3$ steps \\
        Stochastic Jitter     & Bernoulli Delay ($p$) 
        & Delay probability: $0.0$--$0.1$ \\
        Tracking Failure     & Random Replacement 
        & Failure probability: $0.0$--$0.3$ \\
        Biased Noise          & Additive Uniform 
        & Noise: $\mathcal{U}[-1, 1]$\,deg \newline
          Bias:  $\mathcal{U}[-0.1, 0.1]$\,deg \\

        \bottomrule
    \end{tabular}
    }
\end{table}

\subsubsection{Perception Noise Generator} The various noise terms used in the perception noise model and their distribution parameters are listed in Table XIV.

\subsection{Visual Object Representation and Augmentations}

\subsubsection{Pre-Rasterization Augmentations} We outline the general algorithm for applying the pre-rasterization augmentations (\cref{sec:augmentations}) to the Gaussian scenes in Algorithm~1. The parameters used for different augmentation layers are provided in \cref{tab:splat_augmentations}.

\begin{algorithm}[t]
\caption{Pre-Rasterization Augmentation for Gaussians}
\label{alg:cluster_augmentation}
\begin{algorithmic}[1]
\Require
\Statex Gaussian parameters $\mathbf{S} \in \mathbb{R}^{N \times D}$
\Statex Cluster assignments $\mathbf{C} \in \{1,\dots,K\}^N$
\Statex Augmentation probability $p_{\text{aug}}$
\Statex Cluster activation fraction $p_{\text{cluster}}$
\Statex Perturbation range $[\delta_{\min}, \delta_{\max}]$
\Statex Augmentation operator $\oplus \in \{+, \times\}$
\Ensure Augmented Gaussian parameters $\mathbf{S}'$

\State $\mathbf{S}' \leftarrow \mathbf{S}$

\State Sample $p \sim \mathcal{U}(0,1)$
\If{$p > p_{\text{aug}}$}
    \State \Return $\mathbf{S}'$
\EndIf

\For{each cluster $k \in \{1,\dots,K\}$}
    \State Sample $p_k \sim \mathcal{U}(0,1)$
    \If{$p_k > p_{\text{cluster}}$}
        \State \textbf{continue}
    \EndIf

    \State Sample $\delta \sim \mathcal{U}(\delta_{\min}, \delta_{\max})$
    \State $\mathcal{I}_k \leftarrow \{ i \mid C_i = k \}$

    \State $\mathbf{S}'[\mathcal{I}_k] \leftarrow \mathbf{S}'[\mathcal{I}_k] \;\oplus\; \delta$
\EndFor

\State \Return $\mathbf{S}'$
\end{algorithmic}
\end{algorithm}

\subsubsection{Post-process Image Augmentations}
Consistent with prior work~\cite{handa2023dextreme, singh2024dextrah}, our baselines utilize standard post-process image augmentations for data randomization (see \cref{sec:pose_results}). A complete list of these augmentations and parameters is provided in Table XV.

\begin{table}[h]
\centering
\caption{Image Augmentation Operators and Parameters}
\label{tab:post_image_augmentations}
\resizebox{\columnwidth}{!}{%
\begin{tabular}{l c C{2.2cm}}
\toprule
\multicolumn{1}{l}{\textbf{Augmentation}} & \multicolumn{1}{c}{\textbf{Probability}} & \multicolumn{1}{c}{\textbf{Range}} \\
\midrule

\multicolumn{3}{l}{\textbf{\textsc{Photometric Augmentations}}} \\
\addlinespace[0.3em]
Color Jitter & $0.2$ & $[0.8, 1.2]$ \\
Hue Shift & $0.2$ & $[-0.2, 0.2]$ \\
Brightness Scaling & $0.5$ & $[0.5, 1.5]$ \\
Contrast Scaling & $0.5$ & $[0.5, 1.5]$ \\
Gamma & $0.5$ & $[0.5, 1.5]$ \\
Saturation Scaling & $0.5$ & $[0.5, 1.5]$ \\

\addlinespace[0.6em]
\multicolumn{3}{l}{\textbf{\textsc{Sensor and Noise Augmentations}}} \\
\addlinespace[0.3em]
ISO-like Noise & $0.25$ \\
Motion Blur & $0.5$ & Kernel size: $3$--$17$ \\

\addlinespace[0.6em]
\multicolumn{3}{l}{\textbf{\textsc{Blur and Filtering}}} \\
\addlinespace[0.3em]
Box Blur & $0.5$ & Kernel Size: $3$--$5$\\
Binary Opening & $1.0$ & Kernel size: $3$ \\

\bottomrule
\end{tabular}
}
\end{table}

\subsection{Visual Object Pose Estimator Training}

\subsubsection{Network Architecture and Training}
The pose estimator employs a ResNet-34~\cite{He2015DeepRL} backbone, initialized with weights pre-trained on ImageNet. The network receives $120 \times 120$ pixel RGB images, which are normalized and upsampled to $224 \times 224$ pixels to align with the backbone's input size. Feature maps extracted from the final convolutional layer are spatially compressed via adaptive average pooling. These features are then processed by a projection head consisting of a Multi-Layer Perceptron (MLP) with two hidden layers of $512$ and $256$ units, and ReLU activations. We formulate the training as a supervised regression problem, minimizing the Huber loss ($\delta=0.05$) between the predicted and ground-truth keypoint coordinates. The network parameters are optimized using AdamW with a base learning rate of $1 \times 10^{-4}$, employing a schedule that combines linear warmup and cosine annealing.